\documentclass[letterpaper, 10 pt, journal, twoside]{IEEEtran}
\usepackage{xspace} % in order to handle spaces after new commands, see \td below 
\usepackage{amsfonts} % for equations
\usepackage[usenames,dvipsnames]{xcolor} % for colors
\usepackage[]{units} % for units
\usepackage{graphicx} % to rotate figures
\graphicspath{{./includes/}}
\usepackage{hyperref}
\hypersetup{colorlinks=true, citecolor=MidnightBlue, linkcolor=MidnightBlue, urlcolor=MidnightBlue}
\usepackage{cite} % needed to compress references
\newcommand{\boldSubSec}[1]{\noindent\textbf{#1.}}
\newtheorem{remark}{Remark}
\newcommand{\remref}[1]{Remark~\ref{#1}}
\newcommand{\appref}[1]{Appendix~\ref{#1}}
\newcommand{\eref}[1]{(\ref{#1})} % equations
\newcommand{\fref}[1]{Fig.~\ref{#1}} % figures
\newcommand{\sref}[1]{Section~\ref{#1}} % figures
\newcommand{\vref}[1]{Video~{#1}} % figures
\newcommand{\tref}[1]{Table~\ref{#1}} % tables
\newcommand{\Rnum}{\mathbb{R}} % Symbol fo the real numbers set
\newcommand{\criteria}{\gls{fec}\xspace}
\newcommand{\td}{touchdown\xspace}
\newcommand{\lo}{lift-off\xspace}
\newcommand{\nlo}{next lift-off\xspace}
\newcommand{\hipheight}{hip height\xspace}
\newcommand{\fectuple}{T} % the FEC tuple
\newcommand{\fecout}{\mu_\mathrm{safe}} % the FEC tuple
\newcommand{\hmap}{H} % the FEC heightmap
\newcommand{\hheight}{z_h} % hip heights
\newcommand{\bodyvel}{v_b} % body twist
\newcommand{\gaitparams}{\alpha} % gait parameters
\newcommand{\nominal}{p_n} % nominal foothold
\newcommand{\candidate}{p_c} % candidate foothold
\newcommand{\optimal}{p_*} % optimal foothold
\newcommand{\hmapvfa}{H_{\mathrm{vfa}}} % vfa heightmap
\newcommand{\heurtuple}{\fectuple_{\mathrm{vfa}}} % for the nominal foothold
\newcommand{\g}{g(\heurtuple)} % exact foothold evaluation
\newcommand{\ghat}{\hat{g}(\heurtuple)} % estimated foothold evaluation
\newcommand{\heurtuplevpa}{\fectuple_{\mathrm{vpa}}} % for the nominal foothold
\newcommand{\heurtuplevpaR}{\fectuple_{\mathrm{vpa},j}} % for the nominal foothold
\newcommand{\nsf}{\mathrm{n}_\mathrm{sf}}
\newcommand{\setfeasibles}{\mathcal{F}} 
\newcommand{\settuple}{\mathcal{T}} 
\newcommand{\setheights}{\mathcal{Z}} 
\newcommand{\elementfeasibles}{\mathrm{n}_{\mathrm{sf},i}}
\newcommand{\elementtuple}{T_i} 
\newcommand{\elementheights}{z_{h_i}} 
 
\newcommand{\hmapvpa}{H_{\mathrm{vpa}}} % vpa heightmap
\newcommand{\hmapvpaR}{H_{\mathrm{vpa},j}} % vpa heightmap
\newcommand{\gvpa}{g_{\mathrm{vpa}}(\heurtuplevpa)} % exact foothold evaluation
\newcommand{\ghatvpa}{\hat{g}_{\mathrm{vpa}}(\heurtuplevpa)} % exact VPA
\newcommand{\finitesetfeasibles}{\bar{\mathcal{F}}} 
\newcommand{\finitesettuple}{\bar{\mathcal{T}}} 
\newcommand{\finitesetheights}{\bar{\mathcal{Z}}} 
\newcommand{\hipheightsamples}{N_{z_h}}
\newcommand{\rbfn}{\hat{\mathcal{F}}} 
 
\usepackage[nonumberlist, nogroupskip, section=subsection, numberedsection=autolabel]{glossaries} 
\newglossarystyle{mylist}{%
\setglossarystyle{list}% base this style on the list style
}
\setglossarystyle{mylist} % use this style for the glossaries list
\newacronym{tal}{TAL}{Terrain-Aware Locomotion}
\newacronym{to}{TO}{Trajectory Optimization}
\newacronym{drl}{RL}{Reinforcement Learning}
\newacronym{vital}{ViTAL}{Vision-Based Terrain-Aware Locomotion}
\newacronym{vfa}{VFA}{Vision-Based Foothold Adaptation}
\newacronym{vpa}{VPA}{Vision-Based Pose Adaptation}
\newacronym{fec}{FEC}{Foothold Evaluation Criteria}
\newacronym{cnn}{CNN}{Convolutional Neural Network} 
\newacronym{tr}{TR}{Terrain Roughness}
\newacronym{lc}{LC}{Leg Collision}
\newacronym{kfis}{KF}{Kinematic Feasibility}
\newacronym{fc}{FC}{Foot Trajectory Collision}
\newacronym{nsf}{$\mathrm{n_{sf}}$}{Number of Safe Footholds}
\newacronym{feasibles}{$\mathcal{F}$}{Set of Safe Footholds}
\newacronym{rcf}{RCF}{Reactive Controller Framework}
\newacronym{wbc}{WBC}{Whole-Body Controller}
\newacronym{tbr}{TBR}{Terrain-Based Body Reference}
\newacronym{mpc}{MPC}{Model Predictive Control}
\newacronym{hyq}{HyQ}{Hydraulically actuated Quadruped}
\newglossary{fake}{}{}{} % create a fake glossary to only use the entrie but not include them in the glossaries list
\newacronym[type=fake]{hyqreal}{HyQReal}{}
\newacronym[type=fake]{lf}{LF}{Left-Front}
\newacronym[type=fake]{rh}{RH}{Right-Hind}
\newacronym[type=fake]{imu}{IMU}{Inertial Measurement Unit}
\makeglossaries
\title{ViTAL: Vision-Based Terrain-Aware Locomotion\\for Legged Robots}
\author{Shamel Fahmi, Victor Barasuol, Domingo Esteban, Octavio Villarreal, and Claudio Semini
\thanks{%
Manuscript received November 23, 2021; 
accepted  November 7, 2022.
This article was recommended for publication by 
Associate Editor J. Kober and Editor E. Yoshida upon evaluation of the  reviewers' comments.
(\textit{Corresponding author: Shamel Fahmi.})}
\thanks{The authors are with the Dynamic Legged Systems Lab, Istituto Italiano di Tecnologia (IIT), Genova, Italy (email: firstname.lastname@iit.it).}}
\begin{document}
\maketitle
\begin{abstract}
This work is on vision-based planning strategies for legged robots that separate locomotion planning into foothold selection and pose adaptation. 
Current pose adaptation strategies optimize the robot's body pose relative to \textit{given} footholds. 
If these footholds are not reached, the robot may end up in a state with no reachable safe footholds. 
Therefore, we present a~\gls{vital} strategy that consists of novel pose adaptation and foothold selection algorithms. 
\gls{vital} introduces a different paradigm in pose adaptation that does not optimize the body pose relative to given footholds, but the body pose that maximizes the chances of the legs in reaching safe footholds. 
\gls{vital}~plans footholds and poses based on skills that characterize the robot's capabilities and its terrain-awareness.
We use the \unit[90]{kg}~\acrshort{hyq} and \unit[140]{kg}~\acrshort{hyqreal} quadruped robots to validate \gls{vital}, 
and show that they are able to climb various obstacles including 
stairs, gaps, and rough terrains at different speeds and gaits. 
We compare \gls{vital} with a baseline strategy that selects the robot pose based on given selected footholds, and show that \gls{vital} outperforms the baseline. 
\end{abstract}

\section{Introduction}
\begin{figure}\centering
\includegraphics[width=0.98\columnwidth]{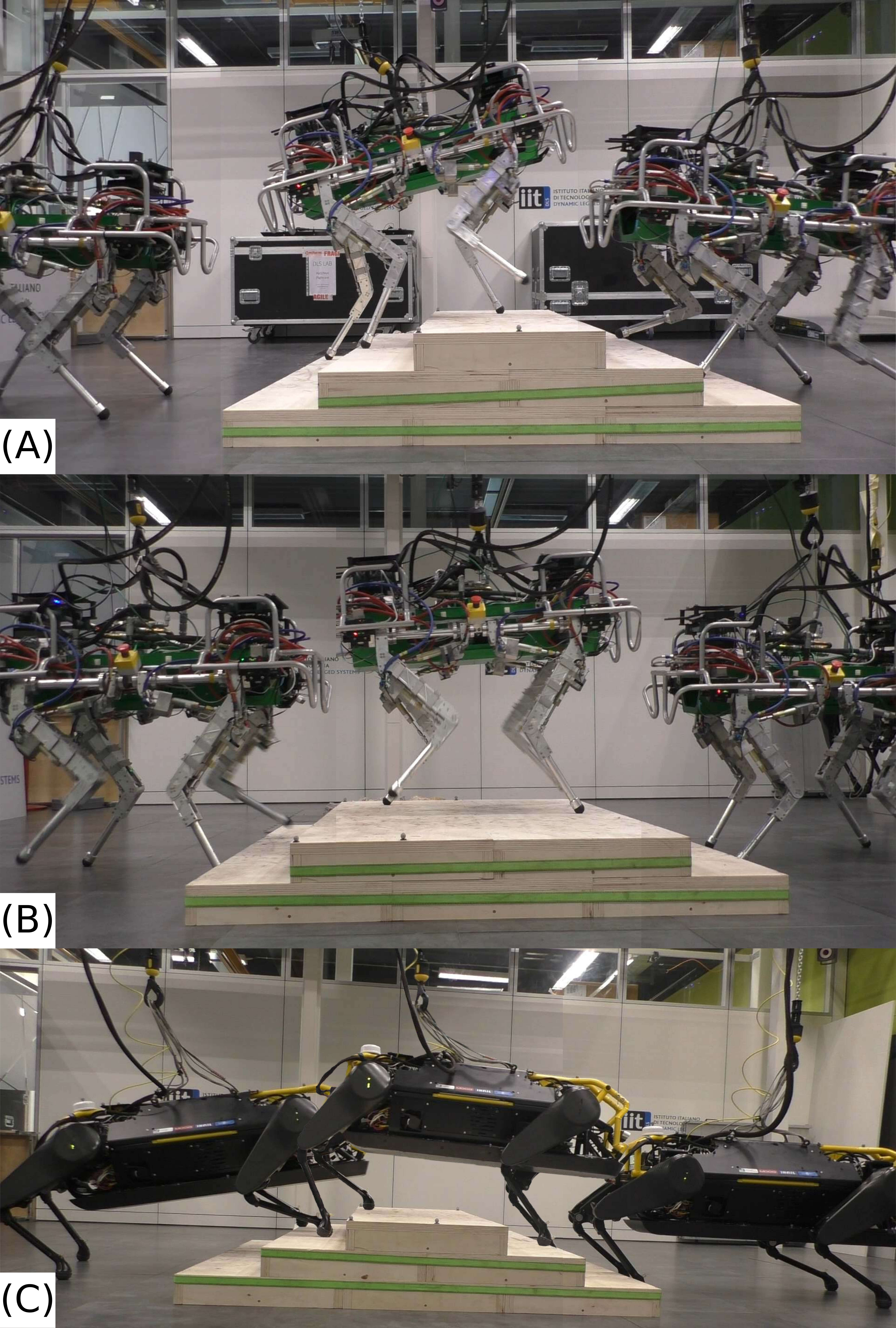}		
\caption{The \acrshort{hyq}~and~\acrshort{hyqreal} quadruped robots climbing stairs using~\gls{vital}.}
\label{fig_1}
\end{figure}

\IEEEPARstart{L}{egged} robots have shown remarkable agile capabilities
in academia~\cite{Lee2021,Yang2020,Semini2019,Katz2019,Bledt2018,Hutter2016}
and industry~\cite{BostonDynamics2021,AgilityRobotics2021,UniTree2021}.
Yet, to accomplish breakthroughs in dynamic whole-body locomotion, 
and to robustly traverse unexplored environments, legged robots have to be \textit{terrain aware}. 
\gls{tal} implies that the robot is capable of taking decisions based on the terrain~\cite{Fahmi2021T}. 
The decisions can be in planning, control, or in state estimation,  
and the terrain may vary in its geometry and physical 
properties~\cite{Buchanan2021,Fahmi2021,Wang2020,Ahmadi2020,Lin2020,Fahmi2020,Paigwar2020,Wellhausen2019,Bosworth2016}.  
\gls{tal}~allows the robot to use its on-board sensors to perceive its surroundings and act accordingly. 
This work is on \textit{vision-based \gls{tal} planning strategies}
that plan the robot's motion (body and feet) based on the terrain information that is acquired using vision (see~\fref{fig_1}).

\subsection{Related Work - Vision-Based Locomotion Planning}\label{sec_related_work_vpa}
Vision-based locomotion planning can either be \textit{coupled} or~\textit{decoupled}.
The coupled approach jointly plans the body pose and footholds in a single algorithm. 
The decoupled approach independently plans the body pose and footholds in separate algorithms.
The challenge in the coupled approach is that it is computationally expensive to solve in real-time. 
Because of this, the decoupled approach tends to be more practical 
since the high-dimensional planning problem is split into multiple low-dimensional problems. 
This also makes the locomotion planning problem more tractable. 
However, this raises an issue with the decoupled approach because the 
plans may conflict with each other since they are planned separately.
Note that both approaches could be solved using optimization, learning, or heuristic methods.

\gls{to} is one way to deal with coupled vision-based locomotion planning.
By casting locomotion planning as an optimal control problem,
\gls{to}~methods can optimize the robot's motion
while taking into account the terrain information~\cite{Fan2021, Melon2021, Ponton2021, Mastalli2020b, Winkler2018b}.
The locomotion planner can generate trajectories
that prevent the robot from slipping or colliding with the terrain
by encoding the terrain's shape and friction properties in the optimization problem~\cite{Winkler2018b}. 
\gls{to}~methods can also include a model of the terrain as a cost map in the optimization problem, 
and generate the robot's trajectories based on that~\cite{Mastalli2020}. 
\gls{to} methods provide guarantees on the optimality and feasibility of the devised motions, 
albeit being computationally expensive; performing these optimizations in real-time is still a challenge.
To overcome this issue,~\gls{to} approaches often implement hierarchical (decoupled) approaches. 
Instead of decoupling the plan into body pose and footholds, the hierarchical approaches 
decouple the plan into short and long-horizon plans~\cite{Brunner2013, Li2020}.
Additionally, 
other work relies on varying the model complexity to overcome the computational issue with~\gls{to}\cite{Li2021}.

\gls{drl} methods mitigate the computational burden of \gls{to} methods by 
training function approximators that learn the locomotion 
plan~\cite{Rudin2022, Yu2022, Miki2021, Gangapurwala2021, Kumar2021, Siekmann2021b, Tsounis2020}.  
Once trained, 
an~\gls{drl} policy can generate body pose and foothold sequences based on proprioceptive and/or visual information. 
Yet, \gls{drl}~methods may require tedious learning (large amounts of data and training time) 
given its high-dimensional state representations.

As explained earlier, decoupled locomotion planning can mitigate
the problems of \gls{to} and \gls{drl} 
by separating the locomotion plan into feet planning 
and body planning~\cite{Kalakrishnan2011,Fankhauser2018,Villarreal2019,Buchanan2020b, Fernbach2018, Tonneau2018}.
Thus, one can develop a more refined and tractable algorithm for every module separately.
In this work, planning the feet motion (foothold locations) is called \textit{foothold selection}, 
and planning the body motion is called \textit{pose adaptation}.

\subsection{Related Work - Foothold Selection and Pose Adaptation}
Foothold selection strategies choose the best footholds
based on the terrain information and the robot's capabilities. 
Early work on foothold selection was presented 
by Kolter~\textit{et~al.}~\cite{Kolter2008} and Kalakrishnan~\textit{et~al.}~\cite{Kalakrishnan2009}
where both approaches relied on motion capture systems 
and an expert user to select (label) the footholds. 
These works were then extended in~\cite{Belter2011} 
using unsupervised learning, on-board sensors,
and considered the terrain information such as 
the terrain roughness (to avoid edges and corners) and friction (to avoid slippage). 
Then, Barasuol~\textit{et~al.}~\cite{Barasuol2015} extended the aforementioned work by selecting footholds
that not only considers the terrain morphology, but also considering leg collisions with the terrain.
Further improvements in foothold selection strategies added other evaluation criteria
such as the robot's kinematic limits.
These strategies use
optimization~\cite{Fankhauser2018,Jenelten2020,Song2021}, 
supervised learning~\cite{Villarreal2019, Belter2019, Esteban2020},
\gls{drl}~\cite{Lee2021,Paigwar2020}, 
or heuristic~\cite{Kim2020} methods.

Similar to foothold selection, pose adaptation strategies 
optimize the robot's body pose based on the terrain information and the robot's capabilities. 
An early work on vision-based pose adaptation was presented in~\cite{Kalakrishnan2011}.
The goal was to find the optimal pose that 
maximizes the reachability of \textit{given} selected footholds, 
avoid collisions with the terrain, 
and maintain static stability. 
The given footholds are based on a foothold selection algorithm 
that considers the terrain geometry.
Another approach was presented in~\cite{Belter2012} that
finds the optimal body elevation and inclination \textit{given}
the selected footholds, and the robot location in the map. 
The pose optimizer 
maximizes different margins that 
increase the kinematic reachability of the legs and static stability,
and avoids terrain collisions. 
This approach was then extended in~\cite{Belter2019b}
with an improved version of the kinematic margins.
A similar approach was presented in~\cite{Fankhauser2018} where
the goal was to find an optimal pose that can maximize 
the reachability of \textit{given} selected footholds. 
The reachability term is accounted for in the cost function 
of the optimizer by penalizing the difference between
the default foothold position and the selected one. 
The work in~\cite{Buchanan2020b} builds on top of 
the pose optimizer of~\cite{Fankhauser2018}
to adapt the pose of the robot in confined spaces using 3D terrain maps. 
This is done using a hierarchical approach 
that first samples body poses that allows the robot to navigate through 
confined spaces, then smooths these poses using a gradient descent method 
that is then augmented with the pose optimizer of~\cite{Fankhauser2018}.
The work presented in~\cite{Villarreal2020} generates vision-based pose references that
also rely on \textit{given} selected footholds
to estimate the orientation of the terrain and send it as a pose reference.
Alongside the orientation reference, 
the body height reference is set at a constant vertical distance (parallel to gravity) 
from the center of the approximated plane that fits through the selected footholds.

The aforementioned pose adaptation strategies 
focus on finding \textit{one} optimal solution 
based on \textit{given} footholds;
footholds have to be first selected and \textit{given} to the optimizer. 
Despite selecting footholds that are safe, 
there are no guarantees on what would happen during execution 
if the footholds are not reached or if the robot deviates from its planned motion.
If any of these cases happen, the robot might end up in a pose where no safe footholds can be reached.
This would in~turn compromise the safety and performance of the robot.
Even if the strategy can re-plan, 
reaching a safe pose might not be possible if the robot is already in an unsafe state.

\subsection{Proposed Approach}
We propose \acrfull{vital}, an online whole-body locomotion planning strategy
that consists of a foothold selection and a pose adaptation algorithm. 
The foothold selection algorithm used in this work is an extension of the \gls{vfa} algorithm 
of the previous work done by Villarreal \textit{et~al.}~\cite{Villarreal2019}
and Esteban \textit{et~al.}~\cite{Esteban2020}. 
Most importantly, 
we propose a novel \gls{vpa} algorithm that
introduces a different paradigm to overcome 
the drawbacks of the state-of-the-art pose adaptation methods. 
\textit{Instead of finding body poses that are optimal for given footholds, 
we propose finding body poses that maximize the chances of reaching safe footholds
in the first place.}
Hence, we are interested in putting the robot in a state 
in which if it deviates from its planned motion, 
the robot remains around a set of footholds that are still 
reachable and safe. 
The notion of safety emerges from \textit{skills} that characterize the robot's capabilities.

\gls{vital} plans footholds and body poses 
by sharing the same robot skills (both for the \gls{vpa} and the \gls{vfa}). 
These skills characterize what the robot is capable of doing.
The skills include, but are not limited to:
the robot's ability to avoid edges, corners, or gaps (\textit{terrain roughness}), 
the robot's ability to remain within the workspace of the legs 
during the swing and stance phases (\textit{kinematic limits}), and
the robot's ability to avoid colliding with the terrain (\textit{leg collision}).
These skills are denoted by \textit{\criteria}.
Evaluating the \criteria is usually computationally expensive. 
Thus, to incorporate the \criteria in \gls{vital}, 
we rely on approximating them with~\glspl{cnn} that are trained via supervised learning.
This allows us to continuously adapt both the footholds and the body pose.
The~\gls{vfa} and the~\gls{vpa} are decoupled and can run at a different update rate. 
However, they are non-hierarchical, they run in parallel, 
and they share the same knowledge of the robot skills (the \criteria). 
By that, we overcome the limitations that result from hierarchical planners as mentioned in~\cite{Buchanan2020b},
where high-level plans may conflict with the low-level ones causing a different robot behavior.

The \gls{vpa} utilizes the \criteria to approximate a function that provides the number of safe footholds for the legs. 
Using this function, we cast a \textit{pose optimizer} which solves a non-linear optimization problem
that maximizes the number of safe footholds for all the legs subject to constraints added to the robot pose. 
The pose optimizer is a key element in the \gls{vpa} since it 
adds safety layers and constraints to the learning part of our approach. 
This makes our approach more tractable 
which mitigates the issues that might arise from end-to-end policies in \gls{drl} methods.

\subsection{Contributions}\label{contribs}
\gls{vital} mitigates the 
above-mentioned conflicts that exist in other decoupled
planners~\cite{Buchanan2020b, Jenelten2020, Tonneau2018, Fankhauser2018}.
This is because both the \gls{vpa} and the \gls{vfa} share the same skills encoded in the \criteria. 
In other words, the \gls{vpa} and the \gls{vfa}
will not plan body poses and footholds that may conflict with each other
because both planners share the same logic. 
In this work, the formulation of the \gls{vpa} allows \gls{vital} to reason about the leg's capabilities and the terrain information.
However, the formulation of the \gls{vpa} could be further augmented by other body-specific skills. 
For instance, the \gls{vpa} could be reformulated to reason about the body collisions with the environment 
similar to the work in~\cite{Tonneau2018, Buchanan2020}. 
The paradigm of the~\criteria can also be further augmented to consider other skills. 
We envision that some skills are best encoded via heuristics while others are well suited 
through optimization. 
For this reason, 
the~\criteria can also handle optimization-based foothold objectives such as the ones in~\cite{Jenelten2020}.

Following the recent impressive results in~\gls{drl}-based locomotion controllers, 
we envision \gls{vital} to be inserted as a module into such control frameworks. 
To elaborate, current \gls{drl}-based locomotion controllers~\cite{Rudin2022, Miki2021, Lee2021, Kumar2021} are of a single network;
the~\gls{drl} framework is a single policy that maps the observations (proprioceptive and exteroceptive) to the actions. 
This may be challenging since it requires careful reward shaping, 
and generalizing to new tasks or different sensors (observations) makes the problem harder~\cite{Green2021}.
For this reason, and similar to Green~\textit{et~al.}~\cite{Green2021}, 
we envision that~\gls{vital} can be utilized as a planner for~\gls{drl} controllers where the~\gls{drl} controller
will act as a reactive controller that then receives guided (planned) commands in a form of optimal poses and footholds from \gls{vital}.

\gls{vital} differs from \gls{to} and optimization-based methods in several aspects. 
The \criteria is designed to independently evaluate every skill (criterion). 
Thus, one criterion can be optimization-based while other could be using logic or heuristics. 
Because of this, \gls{vital} is not restricted by solving an optimization problem that handles all the skills at once. 
Another difference between \gls{vital} and \gls{to} is in the way the body poses are optimized. 
In \gls{to}, the optimization problem optimizes a single pose to follow a certain trajectory. 
The \gls{vpa} in \gls{vital} optimizes for the body poses that maximizes the chances of the legs in reaching safe footholds. 
In other words, the \gls{vpa} finds a body pose that would put the robot in a configuration where the legs have the maximum possible number of safe footholds. 
In fact, this paradigm that the \gls{vpa} of \gls{vital} introduces may be also encoded in \gls{to}.
Additionally, \gls{to} often finds body poses that considers the leg's workspace, 
but to the best of the authors' knowledge, there is no \gls{to} method 
that finds body poses that consider the legs' collision with the terrain, and the feasibility of the swinging legs' trajectory.

\noindent To that end, the \textit{contributions} of this article are
\begin{itemize}
\item \gls{vital}, an online vision-based locomotion planning strategy
that simultaneously plans body poses and footholds based on shared knowledge of robot skills (the~\criteria).
\item 
An extension of our previous work on the \gls{vfa} algorithm for foothold selection
that considers the robot's body twist and the gait parameters.
\item 
A novel pose adaptation algorithm called the \gls{vpa} 
that finds the body pose that maximizes the number of safe footholds for the robot's legs.
\end{itemize}

\section{Foothold Evaluation Criteria (FEC)}\label{sec_heuristics}
The \criteria is main the building block for the \gls{vfa} and the \gls{vpa}. 
The \criteria 
are sets of skills that 
evaluate footholds within heightmaps. 
The skills include 
the robot's ability to assess the terrain's geometry, avoid leg collisions, and avoid reaching kinematic limits.
The \criteria can be 
model-based as in this work and~\cite{Kalakrishnan2011, Villarreal2019}, 
or using optimization techniques as in~\cite{Fankhauser2018, Jenelten2020}. 
The \criteria of this work extends the criteria used in our previous 
work~\cite{Barasuol2015,Villarreal2019,Esteban2020}.

The \criteria takes a tuple~$\fectuple$ as an input,
evaluates it based on multiple criteria, 
and outputs a boolean matrix~$\fecout$.
The input tuple~$\fectuple$ is defined as
\begin{equation}
\fectuple = (\hmap, \hheight, \bodyvel, \gaitparams)
\label{fec_tuble}
\end{equation}
where 
$\hmap\in\Rnum^{h_x\times{h_y}}$ is the heightmap of dimensions $h_x$~and~$h_y$,
$\hheight\in\Rnum$ is the \textit{\hipheight} of the leg (in the world frame), 
${\bodyvel\in\Rnum^6}$ is the base twist,
and $\gaitparams$ are the gait parameters 
(step length, step frequency, duty factor, and time remaining till \td). 
The heightmap $\hmap$ 
is extracted from the terrain elevation map, 
and is oriented with respect to the horizontal frame of the robot~\cite{Barasuol2013}. 
The horizontal frame coincides with the base frame of the robot,
and its $xy$-plane is perpendicular to the gravity vector. 
Each cell (pixel) of~$\hmap$ denotes the terrain height 
that corresponds to the location of this cell in the terrain map. 
Each cell of~$\hmap$ also corresponds to 
a \textit{candidate foothold}~$\candidate\in\Rnum^3$ for the robot.

In this work, we only consider the following \criteria:
\textit{\gls{tr}}, \textit{\gls{lc}}, \textit{\gls{kfis}}, and \textit{\gls{fc}}. 
Each criterion~$C$ 
outputs a boolean matrix~$\mu_C$.
Once all of the criteria are evaluated, the final output~$\fecout$ 
is the element-wise logical AND ($\wedge$)
of all the criteria. 
The output matrix~$\fecout\in\Rnum^{h_x\times{h_y}}$ 
is a boolean matrix with the same size as the input heightmap~$\hmap$.
$\fecout$ indicates the candidate footholds 
(elements in the heightmap $\hmap$) that are \textit{safe}.
An element in the matrix~$\fecout$ that is true, 
corresponds to a candidate foothold $\candidate$
in the heightmap~$\hmap$ that is safe. 
The output of the \criteria is
\begin{equation}
\fecout = 
\mu_\mathrm{TR} \wedge \mu_\mathrm{LC} \wedge \mu_\mathrm{KF} \wedge \mu_\mathrm{FC} .
\label{fec_evaluation}
\end{equation}

\begin{figure*}
\centering
\includegraphics[width=2\columnwidth]{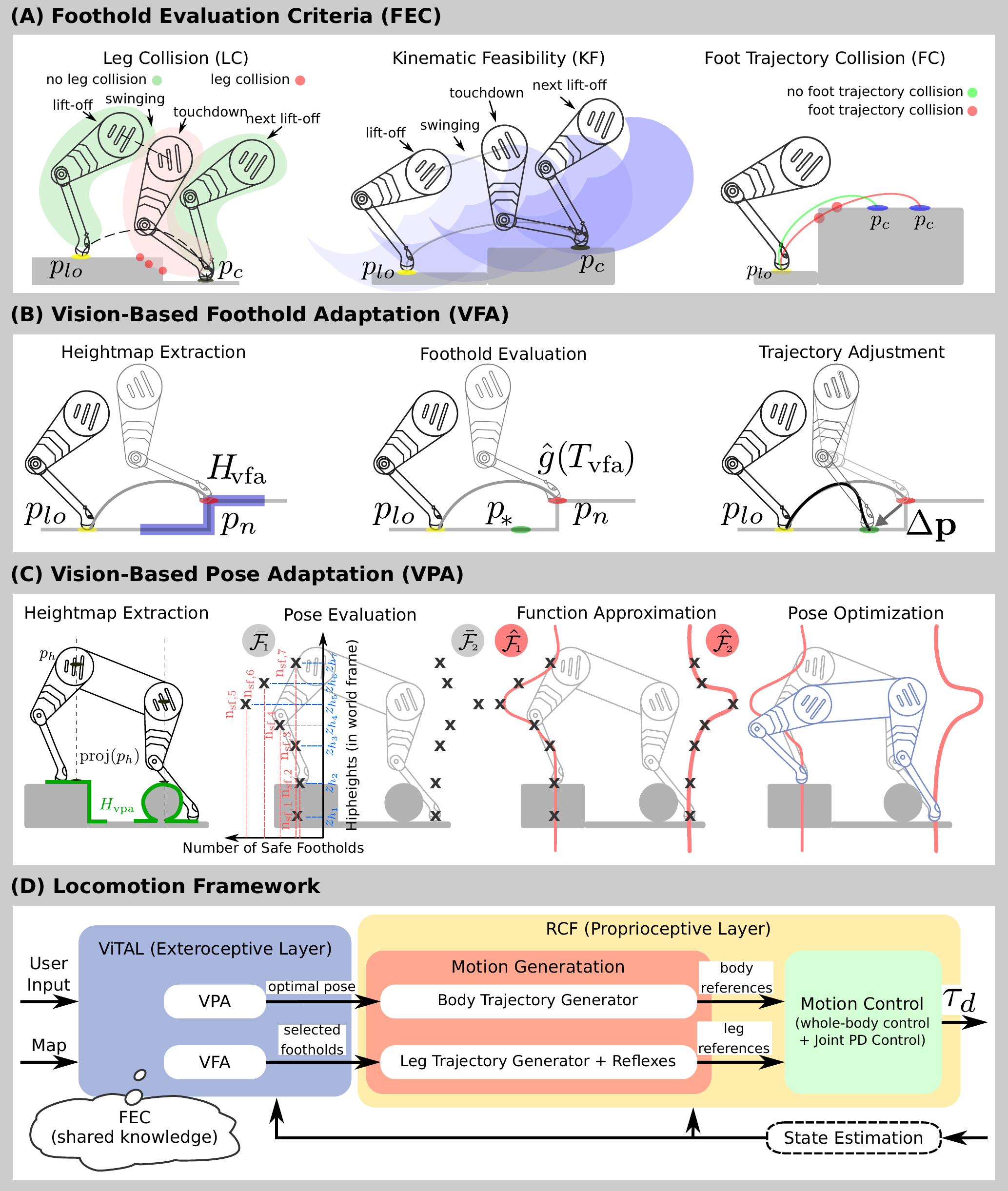}		
\caption
[Overview of \acrshort{vital}.]
{
Overview of \acrshort{vital}. Illustrations are not to scale. 
(A) 
The Foothold Evaluation Criteria (\acrshort{fec}):
Leg Collision~(LC), Kinematic Feasibility~(KF), and Foot Trajectory Collision~(FC). 
(B) 
The Vision-based Foothold Adaptation (VFA) pipeline.
First, we extract the heightmap $\hmapvfa$ around the nominal foothold $\nominal$.
Then, we evaluate the heightmap either using the exact evaluation $\g$ or
using the \acrshort{cnn} as an approximation $\ghat$. 
Once the optimal foothold $\optimal$ is selected, the swing trajectory is adjusted.
(C) 
The Vision-Based Pose Adaptation (VPA) pipeline.
First, we extract the heightmap $\hmapvpa$ for all the legs. The heightmaps are centered around the projection of the leg hip locations. 
Then, we evaluate the \acrshort{fec} to compute $\finitesetfeasibles$ 
for all the hip heights of all the legs (pose evaluation).
Then, we approximate a continuous function $\rbfn$ from  $\finitesetfeasibles$ (function approximation). 
The pose optimizer finds the pose that maximizes $\rbfn$ for all of the legs~(pose optimization).
(D)
Our locomotion framework.
\acrshort{vital} consists of the \acrshort{vpa} and the \acrshort{vfa} algorithms.
Both algorithms rely on the robot skills which we denote by \acrshort{fec}.
$\tau_d$ are the desired joint torques that are sent to the robot.
}
\label{fig_2}
\end{figure*}

An overview of the criteria used in this work is shown in \fref{fig_2}{(A)}
and is detailed below.

\boldSubSec{Terrain Roughness (TR)}
This criterion checks edges or corners in the heightmap that are unsafe for the robot to step on. 
For each candidate foothold $\candidate$ in $\hmap$, we evaluate
the mean and standard deviation of the slope relative to its neighboring footholds, 
and put a threshold that defines whether a $\candidate$ is safe or not. 
Footholds above this threshold are discarded.

\boldSubSec{Leg Collision (LC)}
This criterion selects footholds that do not result in leg collision with the terrain 
during the entire gait cycle (from \lo, during swinging, \td and till the \nlo). 
To do so, we create a bounding region around the leg configuration that corresponds to 
the candidate foothold~$\candidate$ and the current hip location. 
Then, we check if the bounding region collides with the terrain (the heightmap) 
by measuring the closest distance between them.
If this distance is lower than a certain value, then the candidate foothold is discarded.

\boldSubSec{Kinematic Feasibility (KF)}
This criterion selects footholds that are kinematically feasible. 
It checks whether a candidate foothold~$\candidate$ 
will result in a trajectory that remains within 
the workspace of the leg during the entire gait cycle.
To do so, we check if the candidate foothold~$\candidate$
is within the workspace of the leg during \td and \nlo. 
Also, we check if the trajectory of the foot from the \lo position $p_{lo}$
till the \td position at the candidate foothold $\candidate$
is within the workspace of the leg. 
In the initial implementation in \cite{Villarreal2019}, 
this criterion was only evaluated during \td. 
In this work, we consider this criterion during the entire leg step-cycle.

\boldSubSec{Foot Trajectory Collision (FC)}
This criterion 
selects footholds that 
do not result in foot trajectory collision with the terrain.
It checks whether the foot swing trajectory 
corresponding to a candidate foothold $\candidate$
is going to collide with the terrain or not. 
If the swing trajectory collides with the terrain, 
the candidate foothold $\candidate$ is discarded.

\begin{remark}
There are three main sources of uncertainty that can affect
the foothold placement~\cite{Barasuol2015}. 
These sources of uncertainty are due to 
trajectory tracking errors, foothold prediction errors, and drifts in the map. 
To allow for a degree of uncertainty, after computing~$\fecout$, 
candidate footholds that are within a radius of unsafe footholds are also discarded.
This is similar to the erosion operation in image processing.
\label{remark_uncertainty_margin}
\end{remark}

\begin{remark}
The initial implementation of the \criteria in~\cite{Villarreal2019}
only considered the heightmap~$\hmap$ as an input;
the other inputs of the tuple~$\fectuple$ in~\eref{fec_tuble} were kept constant.
This had a few disadvantages that we reported in~\cite{Esteban2020}
where we extended the work of \cite{Villarreal2019} by  
considering the linear body heading velocity. 
In this work, we build upon that by considering
the full body twist~$\bodyvel$ and the gait parameters~$\gaitparams$ as expressed by $\fectuple$ in~\eref{fec_tuble}. 
\label{remark_vfa_differences}
\end{remark}

\section{Vision-Based Foothold Adaptation (VFA)}\label{vfaa}
The \gls{vfa} evaluates the \criteria to select the optimal foothold
for each leg~\cite{Barasuol2015, Villarreal2019, Esteban2020}.
The \gls{vfa} has three main stages as shown in \fref{fig_2}{(B)}:
\textit{heightmap extraction},  \textit{foothold evaluation}, and \textit{trajectory adjustment}.

\boldSubSec{Heightmap Extraction}
Using the current robot states and gait parameters, 
we estimate the \td position of the swinging foot in the world frame as detailed in \cite{Villarreal2019}.
This is denoted as the \textit{nominal foothold} $\nominal\in\Rnum^3$.
Then, we extract a heightmap $\hmapvfa$ that is centered around $\nominal$.

\boldSubSec{Foothold Evaluation}
After extracting the heightmap, 
we compute the \textit{optimal foothold} $\optimal\in\Rnum^3$ for each leg.
We denote this by foothold evaluation which is the mapping 
\begin{equation}
\g: \heurtuple \rightarrow \optimal
\end{equation}
that takes an input tuple~$\heurtuple$ that is defined as
\begin{equation}
\heurtuple =
(\hmapvfa, \hheight, \bodyvel, \gaitparams, \nominal) .
\label{vfa_tuble}
\end{equation}

\begin{figure}[t!]
\centering
\includegraphics[width=\columnwidth]{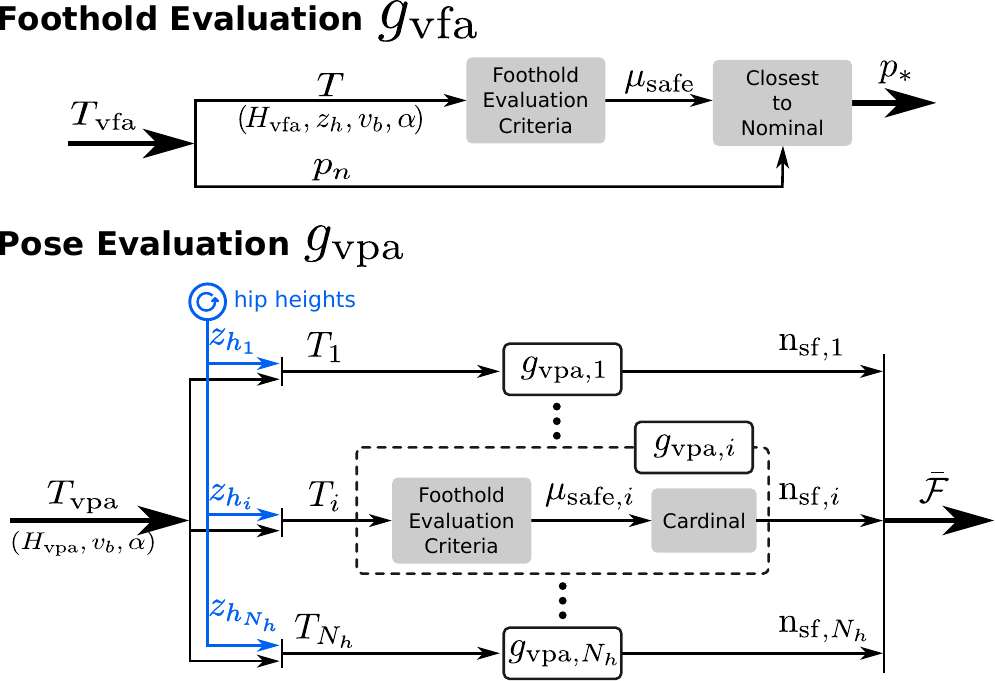}		
\caption{Overview of the foothold evaluation stage in the VFA algorithm, and the pose evaluation stage in the VPA algorithm.}
\label{vfa_footholdeval_vpa_pose_eval}
\end{figure}

Once we evaluate the \criteria in~\eref{fec_evaluation}, 
from all of the safe candidate footholds in $\fecout$, 
we select the optimal foothold~$\optimal$ as the one 
that is \textit{closest to the nominal} foothold $\nominal$.
The aim is to minimize the deviation from the original trajectory 
and thus results in a less disturbed or aggressive motion. 
An overview of the foothold evaluation stage is shown in~\fref{vfa_footholdeval_vpa_pose_eval}
where the tuple~$\fectuple$ of the \criteria in~\eref{fec_tuble} 
is extracted from the \gls{vfa} tuple $\heurtuple$ in~\eref{vfa_tuble}
to compute $\fecout$.
Then, using $\nominal$ and $\fecout$, we extract $\optimal$ as the safe foothold that is closest to $\nominal$.

\boldSubSec{Trajectory Adjustment}
The leg's swinging trajectory is adjusted once~$\optimal$ is computed.

\begin{remark}To compute the foothold evaluation, one can directly apply the exact mapping~$\g$. 
Yet, computing the foothold evaluation leads to 
evaluating the \criteria which is generally computationally expensive. 
Thus, to speed up the computation and to continuously run the \gls{vfa} online, 
we train a~\gls{cnn} to approximate the foothold evaluation~$\ghat$ using supervised learning.
Once trained, the \gls{vfa} can then be executed online using the \gls{cnn}.
The \gls{cnn} architecture of the foothold evaluation is explained in~\appref{cnn_approx}.
\label{remark_cnn_exact}
\end{remark}

\section{Vision-Based Pose Adaptation (VPA)}\label{sec_vpa}
The \gls{vpa} generates pose references that maximize the chances of the legs to reach safe footholds. 
This means that the robot pose has to be aware of what the legs are capable of and adapt accordingly. 
Therefore, the goal of the \gls{vpa} is to adapt the robot pose based on 
the same set of skills in the \criteria used by the~\gls{vfa}.

\subsection{Definitions and Notations}
\boldSubSec{Number of Safe Footholds}
As explained earlier, the \criteria takes a tuple~$\fectuple$ as an input 
and outputs the matrix $\fecout$.
Based on that, let us define the \textit{\gls{nsf}} 
\begin{equation}
\nsf := \mathrm{cardinal}( \{e \in \fecout : e=1   \} )
\label{cardinal}
\end{equation}
as the number of \textit{true} elements in 
the boolean matrix~$\fecout$.

\boldSubSec{Set of Safe Footholds}
Consider a set of tuples~$\settuple$
where each element $\elementtuple \in \settuple$ is a tuple defined as
\begin{equation}
T_i = (H, \elementheights,\bodyvel, \gaitparams)
\label{vpa_eq_2}
\end{equation}
and  $\elementheights \in \setheights$ is a hip height element in 
the set of hip heights~$\setheights$ (in the world frame).
All the tuple elements~$\elementtuple \in \settuple$
share the same heightmap~$\hmap$, body twist~$\bodyvel$ and gait parameters~$\gaitparams$.

Evaluating the \criteria~in~\eref{fec_evaluation} 
for every~$\elementtuple \in \settuple$ that corresponds to~$\elementheights \in \setheights$,
and computing the cardinal~in~\eref{cardinal} yields~$\elementfeasibles$ for every~$\elementtuple$.
This yields the~\textit{\gls{feasibles}}
which is a set containing 
the number of safe footholds $\nsf$ 
that are evaluated based on the \criteria 
given the set of tuples $\settuple$
that corresponds to the set of hip heights $\setheights$
but shares the same heightmap~$\hmap$, body twist~$\bodyvel$ and gait parameters~$\gaitparams$.

\subsection{From the Set of Safe Footholds to Pose Evaluation}\label{pose_evaluation}
The set of safe footholds \gls{feasibles} is one of the building blocks of the \gls{vpa}. 
To compute \gls{feasibles}, we compute the input tuple~$\heurtuplevpa$ 
\begin{equation}
\heurtuplevpa = (\hmapvpa,\bodyvel, \gaitparams)
\label{vpa_tuble}
\end{equation}
that we then augment with the hip heights $\elementheights$ 
in the hip heights set $\setheights$
yielding the set of tuples~$\settuple$. 
Then, we evaluate the \criteria~in~\eref{fec_evaluation} 
for every~$\elementtuple \in \settuple$,
and computing the cardinal~in~\eref{cardinal}.
This can be expressed by the mapping
\begin{equation}
{\gvpa~:~\heurtuplevpa\rightarrow\mathcal{F}}
\label{vpa_eq_3}
\end{equation}
which is referred to as \textit{pose evaluation}.
We can express $\setfeasibles$ as
\begin{equation}
\setfeasibles = 
\{ \elementfeasibles = g_{\mathrm{vpa},i}(\elementtuple)~\forall \elementtuple \in \settuple \}.
\label{vpa_4}
\end{equation}

\begin{remark}
Since $\setheights$ is an infinite continuous set,
so is~$\setfeasibles$
which is not numerically feasible to compute.
Hence, we sample a finite set~$\finitesetheights$ of $\hipheightsamples$ samples of hip heights
that results in a finite set of safe footholds~$\finitesetfeasibles$.
To use the set of safe footholds in an optimization problem, we need a continuous function. 
Thus, after we compute~$\finitesetfeasibles$,  
we estimate a continuous function~$\rbfn$
as explained next.
\label{rem3}
\end{remark}

An overview of the pose evaluation is shown in~\fref{vfa_footholdeval_vpa_pose_eval}
where the tuple~$\heurtuplevpa$ in~\eref{vpa_tuble} is augmented with the 
hip heights $\elementheights$
from~$\finitesetheights$ to construct the \criteria tuples~$\elementtuple$ in~\eref{vpa_eq_2}. 
For every tuple~$\elementtuple$, we evaluate the \criteria using~\eref{fec_evaluation} and 
compute~$\elementfeasibles$ in~\eref{cardinal} using the mapping in~\eref{vpa_eq_3}.
Finally, the set $\finitesetfeasibles$ includes all the elements~$\elementfeasibles$
as in~\eref{vpa_4}.

\subsection{Vision-based Pose Adaptation (VPA) Formulation}\label{sec_vpa_pipeline}
The \gls{vpa} has four main stages as shown in~\fref{fig_2}{(C)}.
First, \textit{heightmap extraction} that is similar to the \gls{vfa}.
Second, \textit{pose evaluation} where we compute $\finitesetfeasibles$. 
Third, \textit{function approximation} where we estimate~$\rbfn$ from~$\finitesetfeasibles$.
Fourth, \textit{pose optimization} where the optimal body pose is computed.

\boldSubSec{Heightmap Extraction}
We extract one heightmap~$\hmapvpa$ per leg
that is centered around the projection of the leg's hip location in the terrain map
($\mathrm{proj.}(p_h)$ instead of~$\nominal$).

\boldSubSec{Pose Evaluation}
After extracting the heightmaps, we compute~$\finitesetfeasibles$
from the mapping in~\eref{vpa_eq_3}
of the pose evaluation.
In the pose evaluation, the \criteria are evaluated 
for all hip heights in~$\finitesetheights$
given the input tuple~$\heurtuplevpa$ as shown in~\fref{vfa_footholdeval_vpa_pose_eval}.

\boldSubSec{Function Approximation}
In this stage, 
we estimate the continuous function~$\rbfn$ from~$\finitesetfeasibles$,
as explained in Remark~\ref{rem3}.
This is done by training a parameterized model
of the inputs $\elementtuple \in \finitesettuple$
and the outputs $\elementfeasibles \in \finitesetfeasibles$.
The result is the function (model)~$\rbfn$
that is parameterized by the model parameters~$w$.
The function approximation is detailed later in this section.

\boldSubSec{Pose Optimization}
Evaluating the~\criteria and approximating it with the function~$\rbfn$,
introduces a metric that represents the possible number of safe footholds for every leg. 
Based on this, the goal of the pose optimizer is to find the optimal pose that
will maximize the number of safe footholds for every leg (maximize~$\rbfn$)
while ensuring robustness.
The pose optimizer is detailed later in this section.

\begin{remark}
Similar to \remref{remark_cnn_exact},
one can directly apply the exact evaluation $\gvpa$ for a given~$\heurtuplevpa$. 
Yet, since this is computationally expensive, we rely 
on approximating the evaluation $\ghatvpa$ using a \gls{cnn}.
In fact,  the learning part is applied to both the pose evaluation 
and the function approximation. 
This means that the pose optimization is running online, 
outside the \gls{cnn}. 
The \gls{cnn} architecture of the pose evaluation is explained in~\appref{cnn_approx}.
\label{remark_cnn_exact_vpa}
\end{remark}

\subsection{Function Approximation}\label{sec_func_approx}
The goal of the function approximation is to approximate 
the set of safe footholds $\setfeasibles$
from the discrete set $\finitesetfeasibles$ computed in the pose evaluation stage.
This is done to provide the pose optimizer with a continuous function.
Given a dataset~$(\finitesetheights, \finitesetfeasibles)$ 
of hip heights~$z_{h_i}\in\finitesetheights$
and number of safe footholds~${\mathrm{n}_{\mathrm{sf},i}\in\finitesetheights}$,
the function approximation estimates a function $\rbfn(z_{h_i}, w)$
that is parameterized by the weights $w$.
Once the weights~$w$ are computed, 
the function estimate~$\rbfn(z_{h_i}, w)$ is then reconstructed and sent to the pose optimizer.

It is important to choose a function~$\rbfn$ that 
can accurately represent the nature of the number of safe footholds. 
The number of safe footholds approaches zero when the hip heights approach~$0$~or~$\infty$.
Thus, we want a function that fades to zero at the extremes
(Gaussian-like functions), and captures any asymmetry or flatness in the distribution. 
Hence, we use radial basis functions of Gaussians. 
With that in mind, we are looking for the weights~$w$ 
\begin{equation}
w = \text{arg}~\text{min}~S(w)
\label{eq_func_aprox_1}
\end{equation}
that minimize the cost~$S(w)$
\begin{equation}
S(w) = \sum_{i=1}^{N_h} (\mathrm{n}_{\mathrm{sf}, i} - \rbfn(z_{h_i}, w))^2
\label{eq_func_aprox_2}
\end{equation}
which is the sum of the squared residuals of 
$\mathrm{n}_{\mathrm{sf}, i}$ and $\rbfn(z_{h_i}, w)$.
$N_h$ is the number of samples (the number of the finite set of hip heights).
The function~$\rbfn(z_{h_i}, w)$ is the regression model 
(the approximation of~$\mathcal{F}$)
that is parameterized by~$w$.
The function~$\rbfn(z_{h_i}, w)$ is the weighted sum of the basis functions
\begin{equation}
\rbfn(z_{h_i}, w) = \sum_{e=1}^{E} w_e \cdot g(z_{h_i}, \Sigma_e, c_e)
\label{eq_func_aprox_3}
\end{equation}
where $w\in\Rnum^E$, and $E$ is the number of basis functions. The basis function is a radial basis function of Gaussian functions%
\begin{equation}
g(z_{h_i}, \Sigma_e, c_e) = \text{exp}(-0.5 (z_{h_i}-c_e)^T \Sigma_e^{-1} (z_{h_i} -c_e))
\label{eq_func_aprox_4}
\end{equation}
where $\Sigma_e$ and $c_e$ are the parameters of the Gaussian function.
Since the function model in~\eref{eq_func_aprox_3} is linear in the parameters, 
the weights of the function approximation can be solved analytically 
using least squares. 
In this work, we
keep the parameters of the Gaussians ($\Sigma$ and $c$) fixed.
Hence, the function 
$\rbfn$ is only parameterized by $w$.
For more information on regression with radial basis functions, 
please refer to \cite{Stulp2015} and~\appref{app_func_approx}.

\subsection{Pose Optimization}\label{sec_po}
The pose optimizer finds the robot's body pose~$u$ 
that maximizes the number of safe footholds for all the legs. 
This is casted as a non-linear optimization problem. 
The notion of safe footholds is provided by the function~$\rbfn(z_h, w)$ that
maps a hip height $z_h$ to a number of safe foothold~\gls{nsf}, 
and is parameterized by $w$.
Since the pose optimizer is solving for the body pose $u$, 
the function~$\rbfn(z_h)$ should be encoded using the body pose rather than the 
hip heights (${\rbfn = \rbfn(z_h(u))}$).
This is done by estimating the hip height as a function of the body pose 
(${z_h = z_h(u)}$) as shown in~\appref{po_appendix}.

\subsection{Single-Horizon Pose Optimization}
The pose optimization problem is formulated as
\begin{eqnarray}
\underset{u=[z_b,\beta, \gamma]}{\text{maximize~}}   &~& 
\mathcal{C} (\rbfn_l( z_{h_l}(z_b,\beta, \gamma) )) ~~ \forall l \in N_l
\label{po_cost_1} \\
\text{subject to} 
&~& u_{\min} \leq u \leq u_{\max} \label{po_u_1} \\
&~& \Delta u_{\min} \leq \Delta u \leq \Delta u_{\max} \label{po_delta_u_1}
\end{eqnarray}
where 
$u=[z_b,\beta, \gamma]\in\Rnum^3$ are the decision variables (robot body pose)
consisting of the robot height, roll and pitch, respectively,
$\mathcal{C}$ is the cost function,
$\rbfn_l$ is $\rbfn$ for every leg $l$ 
where $N_l=4$ is the number of legs, 
$z_{h_l}\in\Rnum$ is the hip height of the leg~$l$,
and
$u_{\min}$ and $u_{\max}$ are the lower and upper bounds
of the decision variables, respectively.
$\Delta u = u - u_{k-1}$ 
is the numerical difference of~$u$ 
where~$u_{k-1}$ is the output of~$u$ at the previous instant,
and
$\Delta u_{\min}$ and $\Delta u_{\max}$ are the lower and upper bounds
of $\Delta u$, respectively.
We can re-write~\eref{po_delta_u_1}~as
\begin{equation}
\Delta u_{\min} + u_{k-1} \leq u \leq \Delta u_{\max} + u_{k-1}. \label{po_delta_u_2}
\end{equation}

The cost function in~\eref{po_cost_1} maximizes~$\rbfn$ for all of the legs.
We designed several types of cost functions as detailed next.
The constraints in~\eref{po_u_1} and~\eref{po_delta_u_1}
ensure that the decision variables and their variations are bounded.

\subsection{Cost Functions}\label{sec_cost_options}
A standard cost function can be the sum of the squares of~$\rbfn_l$ for all 
of the legs
\begin{equation}
\mathcal{C}_{\mathrm{sum}} = \sum_{l=1}^{N_l=4} \Vert \rbfn_l (z_{h_l}) \Vert^2_Q 
\label{eq_cost_option_1}
\end{equation}
where another option could be the product of the squares of~$\rbfn_l$ for all of the legs
\begin{equation}
\mathcal{C}_{\mathrm{prod}} = \prod_{l=1}^{N_l=4} \Vert \rbfn_l (z_{h_l}) \Vert^2_Q .
\label{eq_cost_option_2}
\end{equation}

The key difference between 
an additive cost~$\mathcal{C}_{\mathrm{sum}}$ 
and a multiplicative cost~$\mathcal{C}_{\mathrm{prod}}$  
is that the latter puts equal weighting for each~$\rbfn_l$. 
This is important since we do not want the optimizer
to find a pose that maximizes~$\rbfn$ for one leg 
while compromising the other leg(s).
One can also define the cost 
\begin{equation}
\mathcal{C}_{\mathrm{int}} = \sum_{l=1}^{N_l=4} \Vert 
\int_{z_{h_l}-m}^{z_{h_l}+m} \rbfn_l (z_{h_l}) ~ d z_{h_l} ~
\Vert^2_Q 
\label{eq_cost_option_3}
\end{equation}
which is the \textit{sum of squared integrals}
that can be numerically approximated as
\begin{equation}
\int_{z_{h_l}-m}^{z_{h_l}+m} \rbfn_l (z_{h_l}) ~ d z_{h_l}
\approx
m \cdot ( \rbfn_l (z_{h_l}-m) + \rbfn_l (z_{h_l}+m)) 
\end{equation}
yielding
\begin{equation}
\mathcal{C}_{\mathrm{int}} = \sum_{l=1}^{N_l=4} \Vert    
m \cdot ( \rbfn_l (z_{h_l}-m) + \rbfn_l (z_{h_l}+m) ) 
\Vert^2_Q .
\label{eq_cost_option_4}
\end{equation}

In this cost option, 
we do not find the pose that maximizes~$\rbfn$.
Instead, we want to find the pose that maximizes 
the area around $\rbfn$ that is defined by the margin $m$.
Using $\mathcal{C}_{\mathrm{int}}$ is important since it adds 
robustness in case there is any error in the pose tracking during execution. 
Because of possible tracking errors during execution, 
the robot might end up in the pose~$u^*\pm m$ instead of~$u^*$.
If we use~$\mathcal{C}_{\mathrm{int}}$ as a cost function, 
the optimizer will find poses that maximizes the number of safe footholds
not just for~$u^*$ but within a vicinity of~$m$.
More details on the use of $\mathcal{C}_{\mathrm{int}}$ as a cost function
in the pose optimization of the~\gls{vpa} can be found in~\appref{cint_appendix}.

\subsection{Receding-Horizon Pose Optimization}\label{sec_receding_PO}
Adapting the robot's pose during dynamic locomotion 
requires reasoning about what is ahead of the robot:
the robot should not just consider its current state but also future ones. 
For that, we extend the pose optimizer to consider
the current and future states of the robot in a receding horizon manner. 
To formulate the receding horizon pose optimizer, 
instead of considering~$\rbfn_l~\forall~{l\in N_l}$ in the single horizon case, 
the pose optimizer will consider~$\rbfn_{l,j}~\forall~{l\in N_l},~ {j \in N_h}$
where~$N_h$ is the receding horizon number. 
We compute~$\rbfn_{l,j}$ in the same way explained in the pose evaluation stage. 
More details on computing~$\rbfn_{l,j}$ can be found in~\appref{receding_def}.

The receding horizon pose optimization problem is 
\begin{eqnarray}
\underset{u=[u_1^T, \cdots, u_{N_h}^T]}{\text{maximize}}  &~&
\sum_{j=1}^{N_h} \mathcal{C}_j (\rbfn_{l,j}( z_{h_{l,j}}(u_j) )) \nonumber\\
&+& \sum_{j=1}^{N_h-1} \Vert u_j - u_{j+1} \Vert \nonumber \\
&~&\forall l \in N_l,~ j \in N_h \label{receding_po_cost_1}\\
\text{subject to} &~&
u_{\min} \leq u \leq u_{\max} \\
&~& \Delta u_{\min} \leq \Delta u \leq \Delta u_{\max} 
\label{eq_receding_po}
\end{eqnarray}
where $u=[u_1^T, \cdots, u_j^T, \cdots, u_{N_h}^T]\in\Rnum^{3N_h}$
are the decision variables during the entire receding horizon $N_h$.
Each variable~$u_j = [z_{b,j},\beta_j, \gamma_j]\in\Rnum^3$ is 
the optimal pose of the horizon~$j$.

The first term in~\eref{receding_po_cost_1} 
is the sum of the cost functions~$\mathcal{C}_j$ during the entire horizon~\text{($\forall j \in N_h$)}.
The cost~$\mathcal{C}_j$ can be any of the aforementioned cost functions.
The second term in~\eref{receding_po_cost_1} 
penalizes the deviation between two consecutive  
optimal poses within the receding horizon
($u_j$ and $u_{j+1}$). 
The second term is added so that each optimal pose $u_j$
is also taking into account the optimal pose 
of the upcoming sequence~$u_{j+1}$ (to connect the solutions in a smooth way).
Similar to the single horizon pose optimizer, 
$u_{\min}$ and $u_{\max}$ are the lower and upper bounds
of the decision variables, respectively.
Furthermore,~$\Delta u$ denotes the numerical difference of $u$,
while~$\Delta u_{\min}$ and $\Delta u_{\max}$ are the lower and upper bounds
of $\Delta u$, respectively.
Note that the constraints of the single horizon and the receding horizon 
are of different dimensions.

\section{System Overview}\label{sec_sys_over}
Our locomotion framework that is shown in~\fref{fig_2}(D) is based on the \gls{rcf}~\cite{Barasuol2013}.
\gls{vital} complements the \gls{rcf} with an exteroceptive terrain-aware layer composed of the \gls{vfa} and the \gls{vpa}. 
\gls{vital} takes the robot states, the terrain map and user commands as inputs, 
and sends out the selected footholds and body pose to the \gls{rcf} (perceptive) layer.
The \gls{rcf} takes the robot states and the references from \gls{vital}, 
and uses them inside a motion generation and a motion control block. 
The motion generation block generates the trajectories of the leg and the body, 
and adjusts them with the reflexes from~\cite{Barasuol2013,Focchi2013}. 
The legs and body references from the motion generation block are sent to the motion control block.
The motion control block consists of a \gls{wbc}~\cite{Fahmi2019} 
that generates desired torques that are tracked via a low-level torque
controller~\cite{Boaventura2015}, and sent to the robot's joints.
The framework also includes a state estimation block that feeds 
back the robot states to each of the aforementioned layers~\cite{Camurri2017}. 
More implementation details on~\acrshort{vital} and the entire framework is in~\appref{misc_details}.

We demonstrate \gls{vital} on 
the \unit[90]{kg} \acrshort{hyq} and the \unit[140]{kg} \acrshort{hyqreal} quadruped robots.
Each leg of the two robots has 3~degrees of freedom (3~actuated joints). 
The torques and angles of the 12~joints of both robots are directly measured. 
The bodies of \acrshort{hyq} and \acrshort{hyqreal} have a tactical-grade~\gls{imu} (KVH 1775). 
More information on \acrshort{hyq} and \acrshort{hyqreal}
can be found in~\cite{Semini2011}, and~\cite{Semini2019} respectively.

We noticed a significant drift in the states of the robots in experiment. 
To tackle this issue, the state estimator fused the data from a motion capture system and the \gls{imu}.
This reduced the drift in the base states of the robots albeit not eliminating it completely. 
Improving the state estimation is an ongoing work and is out of the scope of this article.
We used the grid map interface~\cite{Fankhauser2016} to get the terrain map in simulation. 
Due to the issues with state estimation on the real robots, 
we constructed the grid map before the experiments, and used the motion capture system to locate
the map with respect to the robot.

\section{Results}\label{sec_results_vital}
We evaluate~\gls{vital} on~\acrshort{hyq} and~\acrshort{hyqreal}.
We consider all the \criteria mentioned earlier for the~\gls{vfa} and the~\gls{vpa}.
We use the receding horizon pose optimizer of~\eref{eq_receding_po}
and the sum of squared integral of~\eref{eq_cost_option_4}.
We choose stair climbing as an application for~\gls{vital}. 
Climbing stairs is challenging for~\acrshort{hyq} 
due to its limited leg workspace in the sagittal plane.
Videos associated with the upcoming results can be found in the supplementary materials and~\cite{Video}. 
Finally, an analysis of the accuracy of the~\glspl{cnn} and the computational time of~\gls{vital}
can be found in~\appref{est_acc} and~\appref{comp_anal}, respectively.

\subsection{Climbing Stairs (Simulation)}\label{sec_sim1}
\begin{figure}
\centering
\includegraphics[width=\columnwidth]{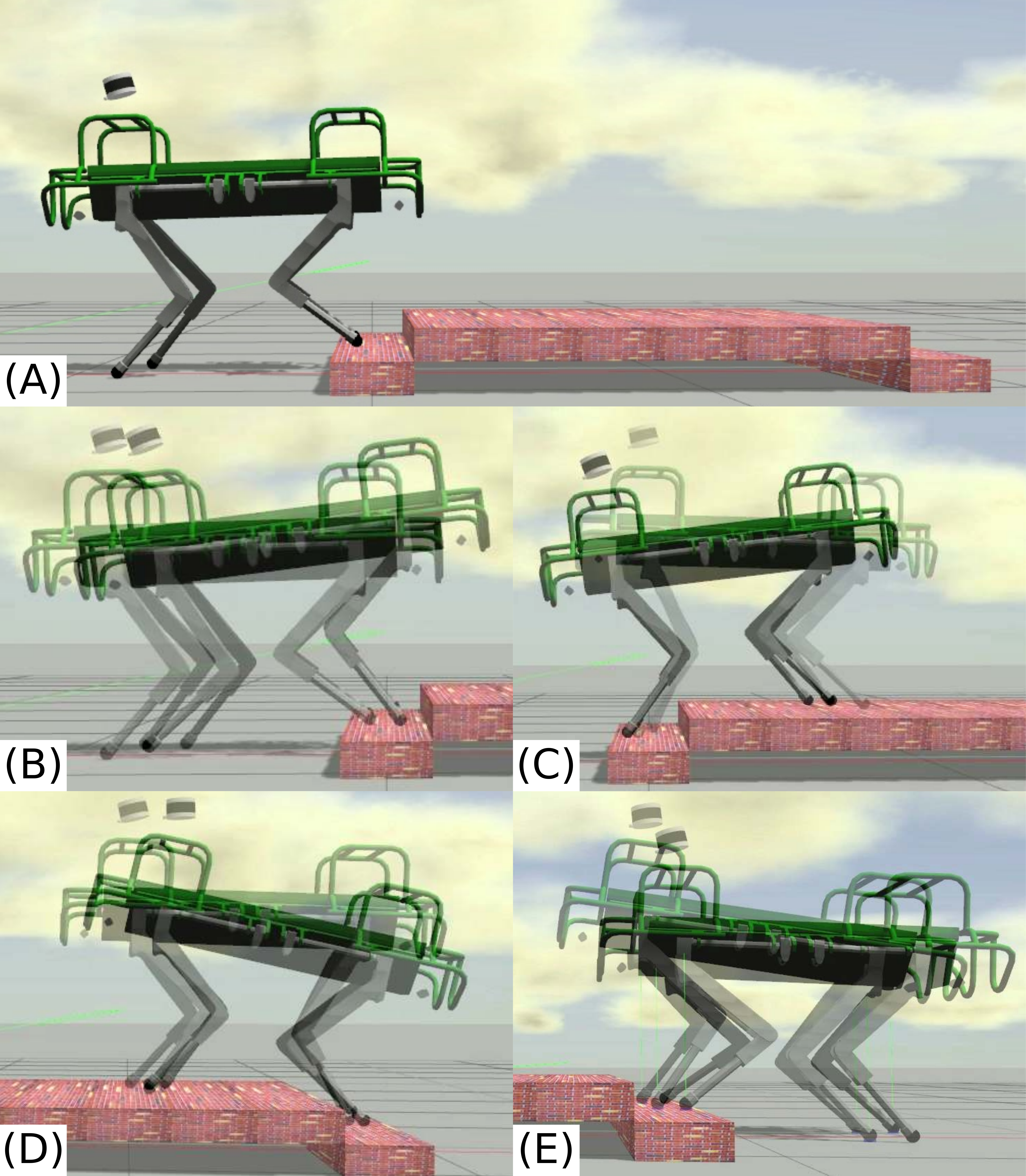}		
\caption
[\acrshort{hyq} climbing stairs in simulation.]
{
\acrshort{hyq} climbing stairs in simulation.
(A)~The full scenario.
(B)~The~robot~pitches up to allow for safe footholds for the front legs.
(C)~The~robot~lifts up the hind hips to avoid hind leg collisions with the step. 
(D)~The~robot~pitches down to allow for safe footholds for the front legs. 
(E)~The~robot~lowers the hind hips to allow for safe footholds for the legs when stepping down.
}
\label{fig_3}
\end{figure}

\begin{figure}
\centering
\includegraphics[width=\columnwidth]{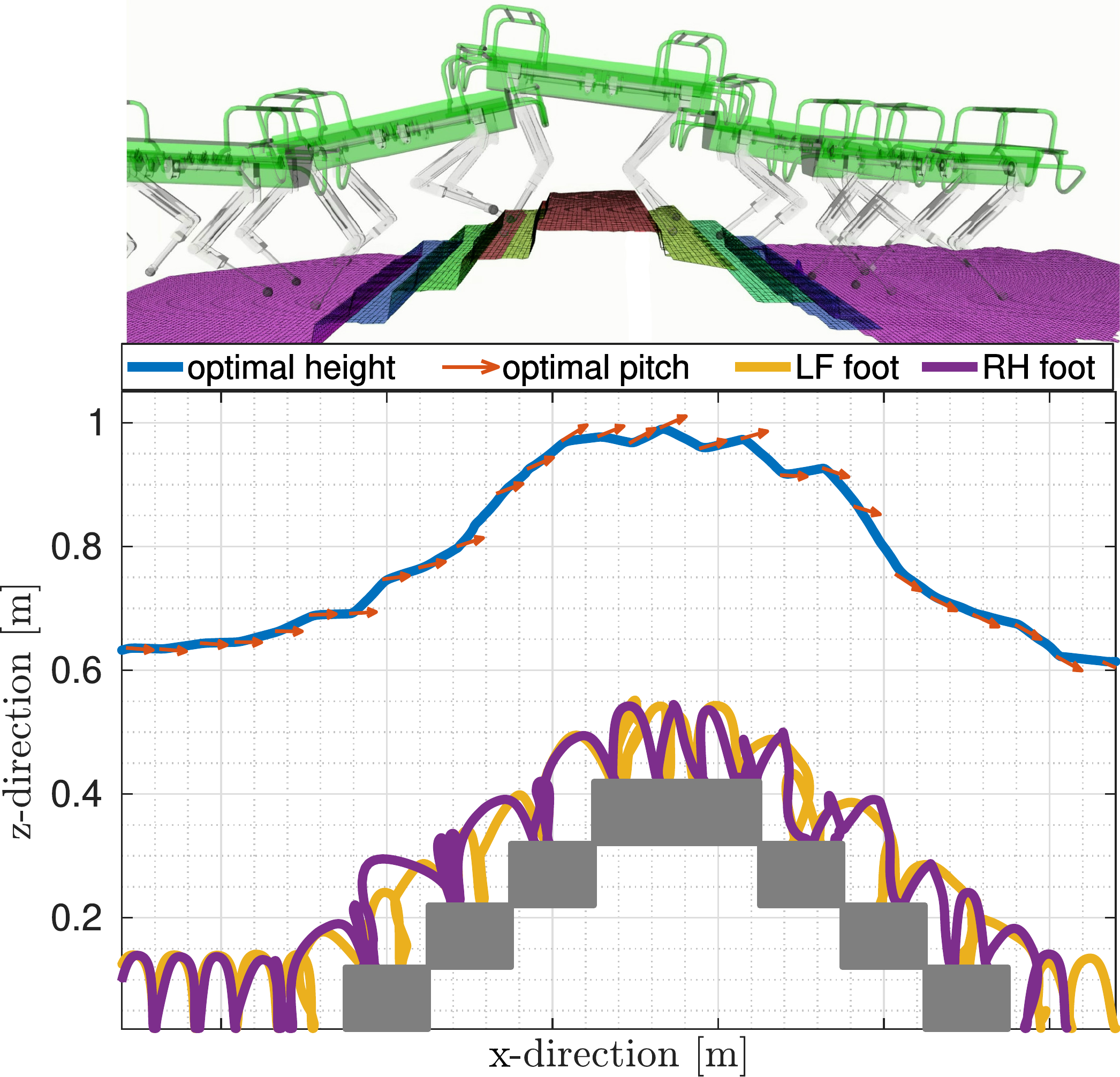}		
\caption[Climbing Stairs: A More Complex Scenario.]{Climbing Stairs: A More Complex Scenario. 	
Top: Overlayed screenshots of \acrshort{hyq} climbing stairs.
Bottom: the optimal height and corresponding pitch (presented by the arrows) 
and the foot trajectories of \acrshort{lf} and \acrshort{rh} legs.}
\label{fig_4}
\end{figure}

We carried out multiple simulations where~\acrshort{hyq} 
is climbing the stairs shown in~\fref{fig_3}.
Each step has a rise of~\unit[10]{cm}, and a go of~\unit[25]{cm}.
\acrshort{hyq} is commanded to trot with a desired forward velocity 
of~\unit[0.2]{m/s} using the~\gls{vpa} and the~\gls{vfa}.
Figure~\ref{fig_3} shows screenshots of one simulation run,
and~\vref{1} shows three simulation runs.

Figure~\ref{fig_3} shows 
the ability of the \gls{vpa} in adapting the robot pose to increase the chances
of the legs to succeed in finding a safe foothold. 
In~\fref{fig_3}(B), 
\acrshort{hyq} raised its body and pitched upwards
so that the front hips are raised 
to increase the workspace of the front legs when stepping up. 
In~\fref{fig_3}(C), 
\acrshort{hyq} raised its body and pitched downwards 
so that the hind hips are raised.
This is done for two reasons. 
First, 
to have a larger clearance between the hind legs and the obstacle, 
and thus avoiding leg collision with the edge of the stairs.
Second,
to increase the workspace of the hind legs when stepping up, 
and thus avoiding reaching the workspace limits and
collisions along the foot swing trajectory.
In~\fref{fig_3}(D), 
\acrshort{hyq} lowered its body and pitched downwards
so that the front hips are lowered. 
This is done for two reasons:
First, 
to increase the workspace of the front legs when stepping down, 
and thus avoiding reaching the workspace limits. 
Second, 
to have a larger clearance between the front legs and the obstacle, 
and thus avoiding leg collisions. 
In~\fref{fig_3}(E), 
\acrshort{hyq} lowered its hind hips
to increase the hind legs' workspace when stepping down.

Throughout these simulations, the robot continuously adapted its body pose and its feet
to find the best trade-off between increasing the kinematic feasibility, 
and avoiding trajectory and leg collision. 
This can be seen in~\vref{1}
where the robot's legs and the corresponding feet trajectories
never collided with the terrain. 
The robot took multiple steps around the same foot location before stepping over an obstacle.
The reason behind this is that the robot 
waited for the \gls{vpa} to change the pose and allow for safe footholds,
and then the \gls{vfa} took the decision of stepping over the obstacle.

We carried out another scenario where \acrshort{hyq} is climbing the stairs setup in~\fref{fig_4}
where each step has a rise of~\unit[10]{cm}, and a go of~\unit[25]{cm}.
\acrshort{hyq} is commanded to trot with a desired forward velocity of $0.2$\unit{m/s} using~\acrshort{vital}.
The results are reported in~\fref{fig_4} and~\vref{2}. 
Figure~\ref{fig_4} shows the robot's height and pitch based on the \acrshort{vpa}, 
and the corresponding feet trajectories of the~\acrshort{lf} leg and the~\acrshort{rh} leg based on the \acrshort{vfa}.
\acrshort{hyq}'s behavior was similar to the previous section:
it accomplished the task without collisions or reaching workspace limits.

\subsection{Climbing Stairs (Experiments)}\label{sec_fig_7}
\begin{figure*}
\centering
\includegraphics[width=\textwidth]{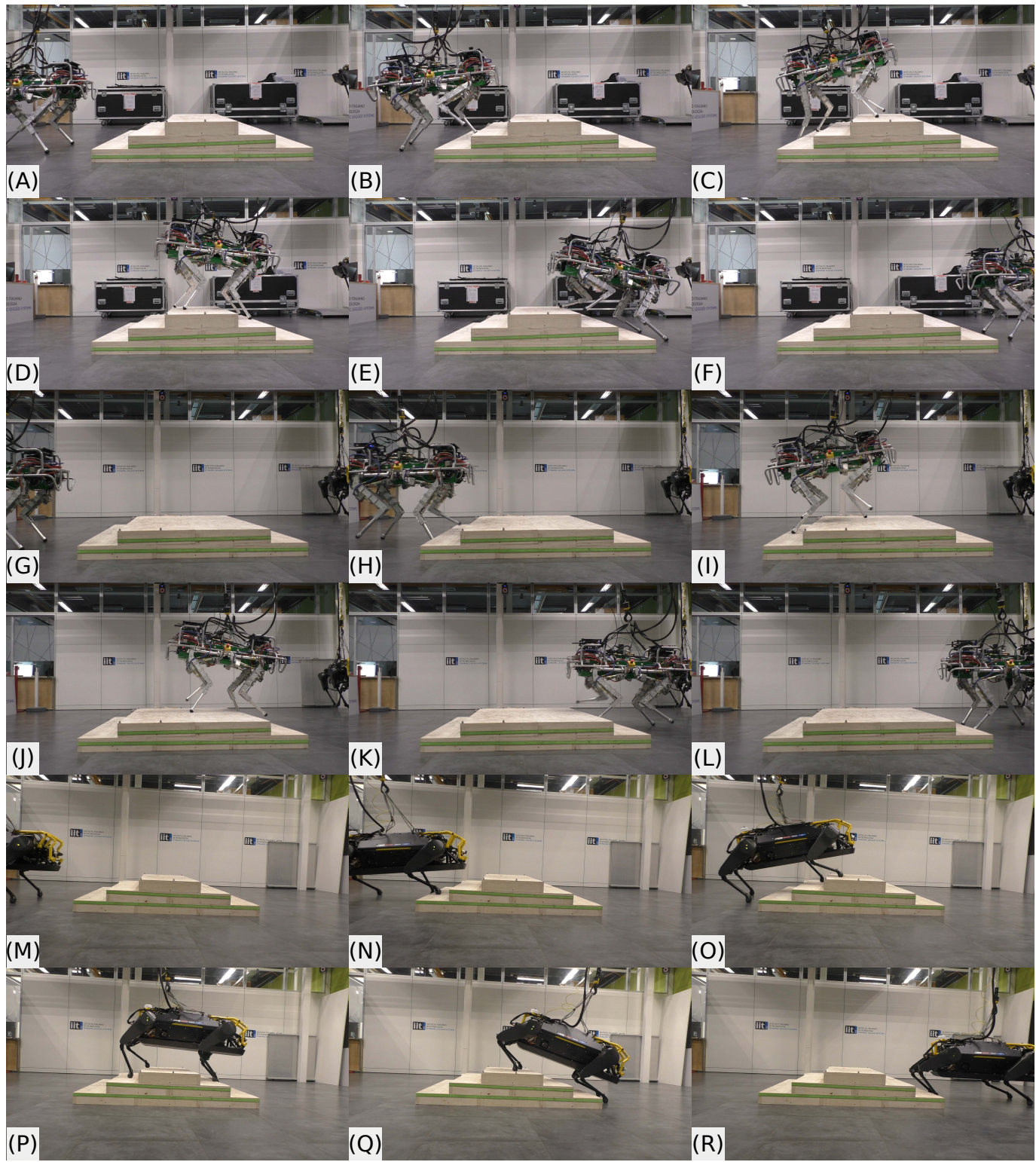}		
\caption
[\acrshort{hyq} and \acrshort{hyqreal} climbing stairs in experiment.]
{\acrshort{hyq} and \acrshort{hyqreal} climbing stairs in experiment.
(A-F)~\acrshort{hyq} crawling over with~\unit[0.1]{m/s} commanded forward velocity.
(G-L)~\acrshort{hyq} trotting over with~\unit[0.25]{m/s} commanded forward velocity.
(M-R)~\acrshort{hyqreal} crawling over with~\unit[0.2]{m/s} commanded forward velocity.
}
\label{fig_6}
\end{figure*} 
\begin{figure}[t!]
\centering
\includegraphics[width=\columnwidth]{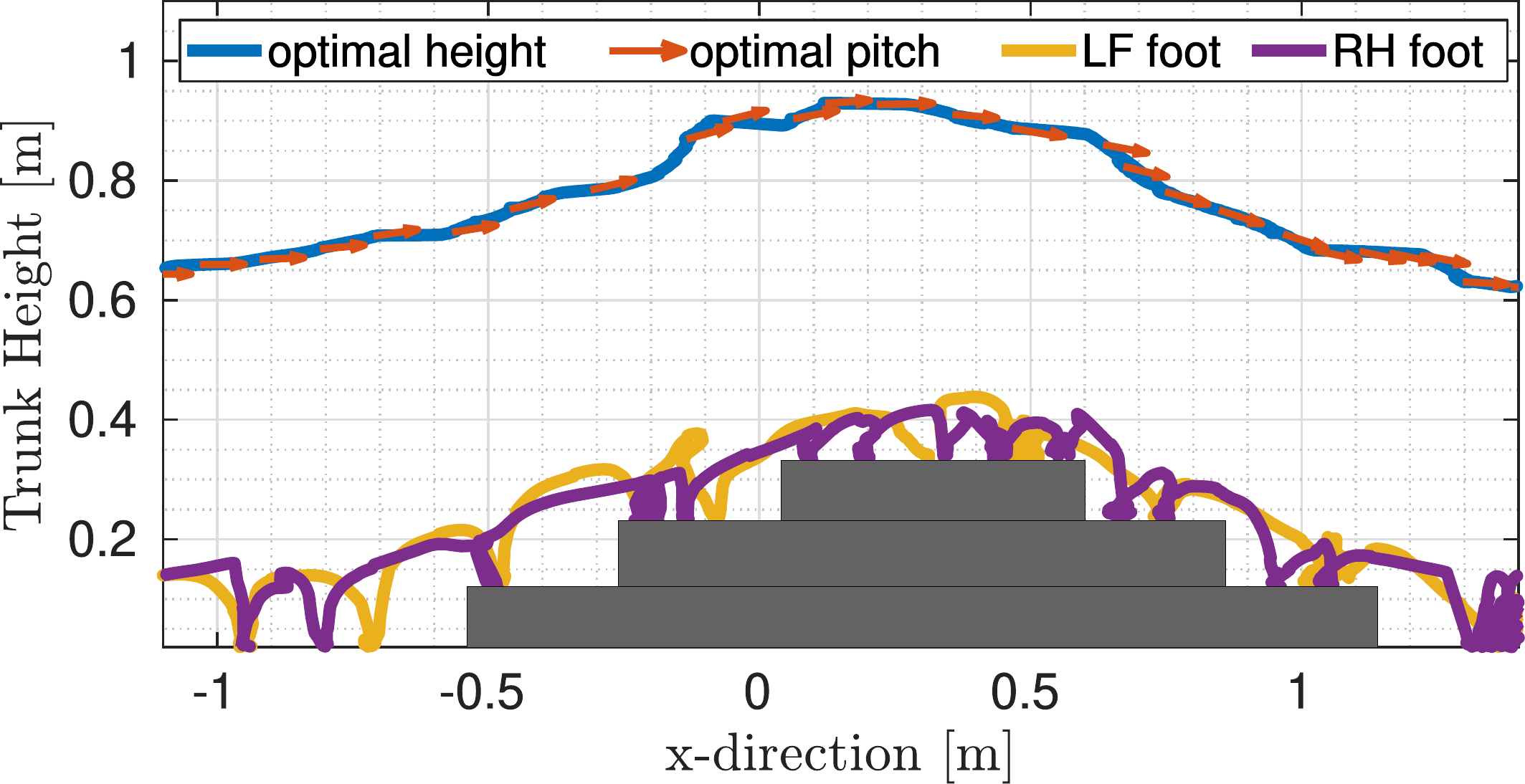}		
\caption
[\acrshort{hyq} climbing stairs in experiment.]
{
\acrshort{hyq} climbing stairs in experiment.
The figure shows the optimal height and corresponding pitch (presented by the arrows) 
based on the \acrshort{vpa}, 
and the foot trajectories of the~\acrshort{lf} and~\acrshort{rh} legs
based on the \acrshort{vfa}.
}
\label{fig_7}
\end{figure} 

To validate \gls{vital} in experiments, 
we created three sets of experiments using
the setups shown in~\fref{fig_1}. 
Each step has a rise of~\unit[10]{cm}, and a go of~\unit[28]{cm}.

In the first set of experiments, 
\acrshort{hyq} is commanded to crawl over the setups in~\fref{fig_1}(A,B)
with a desired forward velocity of~\unit[0.1]{m/s} using the \gls{vpa} and the \gls{vfa}.
\text{Figures~\ref{fig_6}(A-F)} show screenshots of one trial.
\vref{3} shows \acrshort{hyq} climbing back and forth the setup in~\fref{fig_1}(B) five times. 
\vref{4} shows \acrshort{hyq} climbing the~\fref{fig_1}(A) setup, which is reported in~\fref{fig_7}. 
Figure~\ref{fig_7} shows the robot's height and pitch based on the \gls{vpa}, 
and the corresponding feet trajectories of the \gls{lf} leg and the \gls{rh} leg based on the \gls{vfa}.
This set of experiments confirms that~\gls{vital} is effective on the real platform.
The robot managed to accomplish the task without collisions or reaching workspace limits. 

In the second set of experiments, \acrshort{hyq} is commanded to trot 
over the setup in~\fref{fig_1}(B) with a desired forward velocity of~\unit[0.25]{m/s} using the \gls{vpa} and the \gls{vfa}.
Figures~\ref{fig_6}(\text{G-L}) show separate screenshots of this trial.
\vref{5} shows three trials of \acrshort{hyq} climbing the same setup. 
This set of experiments shows that~\gls{vital} can handle different gaits.

Finally, in the third set of experiments,
\acrshort{hyqreal} is commanded to crawl 
over the setup in~\fref{fig_1}(C) with a desired forward velocity of~\unit[0.2]{m/s} 
using the~\gls{vpa} and the~\gls{vfa}.
The results are reported in~\fref{fig_6}(M-R) that show screenshots of this trial.
\vref{6} shows \acrshort{hyqreal} climbing this stair setup (\fref{fig_1}(C)). 
This set of experiments shows that~\gls{vital} can work on different legged platforms.

\begin{table}[t!]
\centering 
\caption{Mean Absolute Tracking Errors of the Body Pitch~$\beta$ and Height~$z_b$
of \acrshort{hyq} \& \acrshort{hyqreal} using \acrshort{vital} in the Experiments 
detailed in \sref{sec_fig_7} and in~\fref{fig_1}.\label{tab_mae_exp}}
\renewcommand{\arraystretch}{1.25}
\begin{tabular}{lcc}
\hline \hline
Description                                              &  $\beta$~[deg] & $z_b$~[cm] \\ \hline
Exp. (A): \acrshort{hyq} crawling at \unit[0.1]{m/s}     &  1.0           & 1.5 \\
Exp. (B): \acrshort{hyq} trotting at \unit[0.2]{m/s}     &  1.0           & 2.3 \\
Exp. (C): \acrshort{hyqreal} crawling at \unit[0.2]{m/s} &  1.0           & 4.2 \\
\hline \hline
\end{tabular}
\end{table}

The tracking performance of these three sets of experiments are shown in~\tref{tab_mae_exp}.
The table shows the mean absolute tracking errors of 
the body pitch~$\beta$ and height~$z_b$ of~\acrshort{hyq} and~\acrshort{hyqreal}.

\subsection{Climbing Stairs with Different Forward Velocities}\label{sec_sim_4}
\begin{figure}[t!]
\centering
\includegraphics[width=\columnwidth]{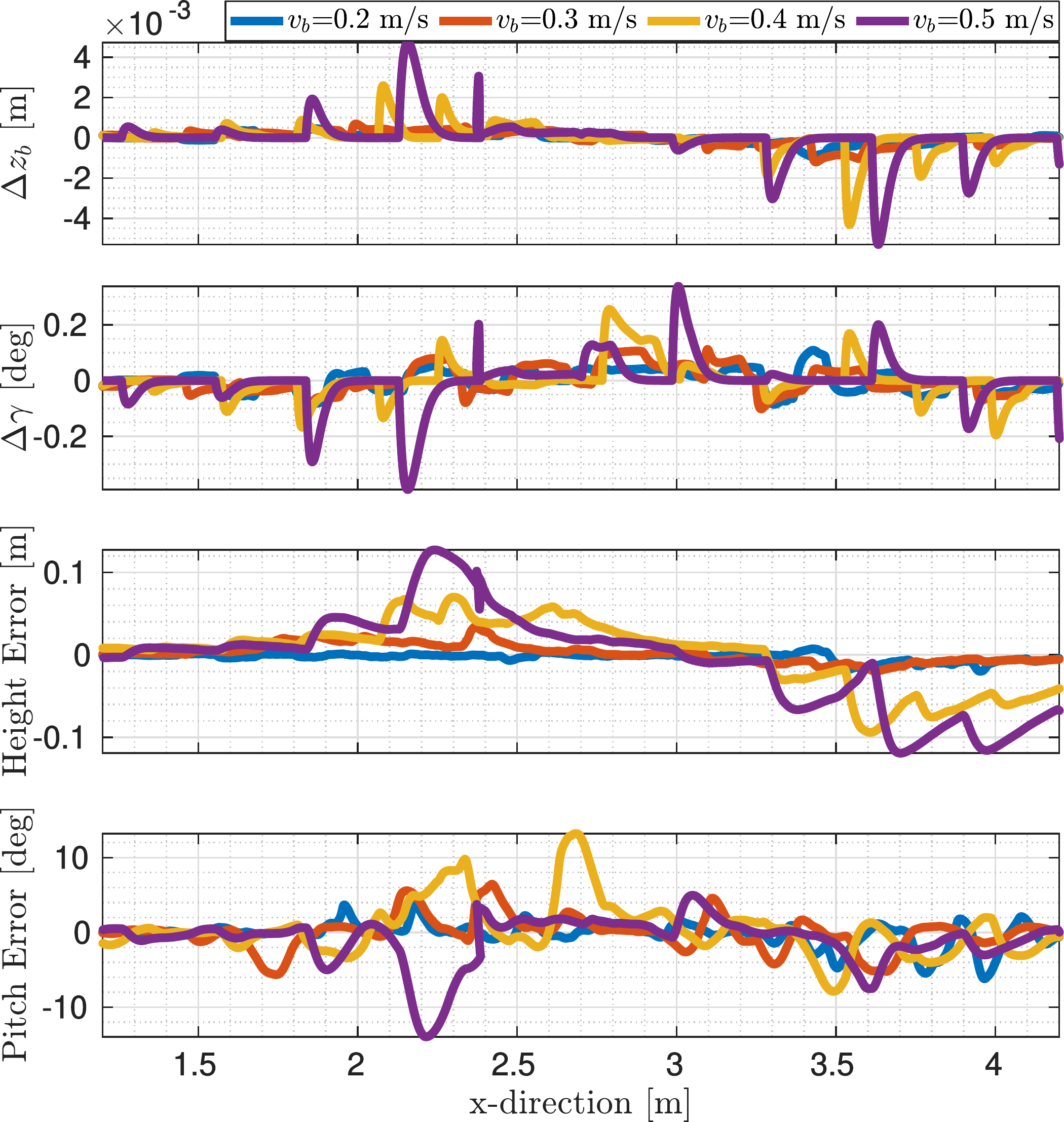}		
\caption
[\acrshort{hyq}'s Performance under Different Commanded Velocities]
{\acrshort{hyq}'s performance using~\gls{vital} under different commanded velocities.
The top two plots show the numerical difference of the body height and pitch ($\Delta z_b$ and $\Delta \gamma$), 
and the bottom two plots show the tracking errors of the body height and pitch.}
\label{fig_8}
\end{figure} 

\begin{table}[t!]
\centering 
\caption{Mean Absolute Tracking Errors of the Body Pitch~$\beta$ and Height~$z_b$
of  \acrshort{hyq} using \acrshort{vital} 
in Simulation with Different Forward Velocities
as Detailed in~\sref{sec_sim_4} and~\fref{fig_8}.\label{tab_mae_sim}}
\renewcommand{\arraystretch}{1.25}
\begin{tabular}{lcc}
\hline \hline
Description & $\beta$~[deg] & $z_b$~[cm]   \\ \hline
Sim. (A): \acrshort{hyq} trotting at \unit[0.2]{m/s} &  0.7 & 0.3 \\
Sim. (B): \acrshort{hyq} trotting at \unit[0.3]{m/s} &  1.1 & 0.7 \\
Sim. (C): \acrshort{hyq} trotting at \unit[0.4]{m/s} &  2.3 & 2.6 \\
Sim. (D): \acrshort{hyq} trotting at \unit[0.5]{m/s} &  1.6 & 3.8 \\
\hline \hline
\end{tabular}
\end{table}

We evaluate the performance of \acrshort{hyq}
under different commanded velocities using~\gls{vital}. 
We carried out a series of simulations using the stairs setup shown in~\fref{fig_4}. 
\acrshort{hyq} is commanded to trot at four different forward velocities:
\unit[0.2]{m/s}, \unit[0.3]{m/s}, \unit[0.4]{m/s}, and~\unit[0.5]{m/s}.
The results are reported in~\fref{fig_8} and in~\vref{7}.
Additionally, 
the tracking performance of these series of simulations
at the four different forward velocities are shown in~\tref{tab_mae_sim}.
The table shows the mean absolute tracking errors of  
the body pitch~$\beta$ and height~$z_b$
of~\acrshort{hyq} at the corresponding commanded forward velocity.
Figure~\ref{fig_8} shows the numerical differences~$\Delta z_b$ and~$\Delta\gamma$, 
and the tracking errors of the body height and pitch, respectively. 

\acrshort{hyq} was able to climb the stairs terrain under different commanded velocities.
However, as the commanded velocity increases, 
\acrshort{hyq} started having faster (abrupt) changes in the body pose
as shown in the top two plots of~\fref{fig_8}. 
As a result, 
the height and pitch tracking errors
increase proportionally to the commanded speed
as shown in the bottom two plots of~\fref{fig_8}.  
This can also be seen in~\tref{tab_mae_sim} where the mean absolute tracking errors
of the pitch and height increase proportionally to the commanded speed.

Similarly to~\acrshort{hyq}, we evaluate \gls{vital} on \acrshort{hyqreal} 
and commanded it to trot with five different forward velocities:
\unit[0.2]{m/s}, \unit[0.3]{m/s}, \unit[0.4]{m/s}, \unit[0.5]{m/s}, and~\unit[0.75]{m/s}.
We report this simulation in~\vref{8}
where we show that \gls{vital} is robot independent. 
Yet, since the workspace of \acrshort{hyqreal} is larger than \acrshort{hyq}, 
this scenario was more feasible to traverse for \acrshort{hyqreal}.
Thus, \acrshort{hyqreal} was able to reach a higher commanded velocity than 
the ones reported for \acrshort{hyq}.

\subsection{Comparing the \acrshort{vpa} with a Baseline (Experiments)}\label{tbr_result}
\begin{figure}[t!]
\centering
\includegraphics[width=\columnwidth]{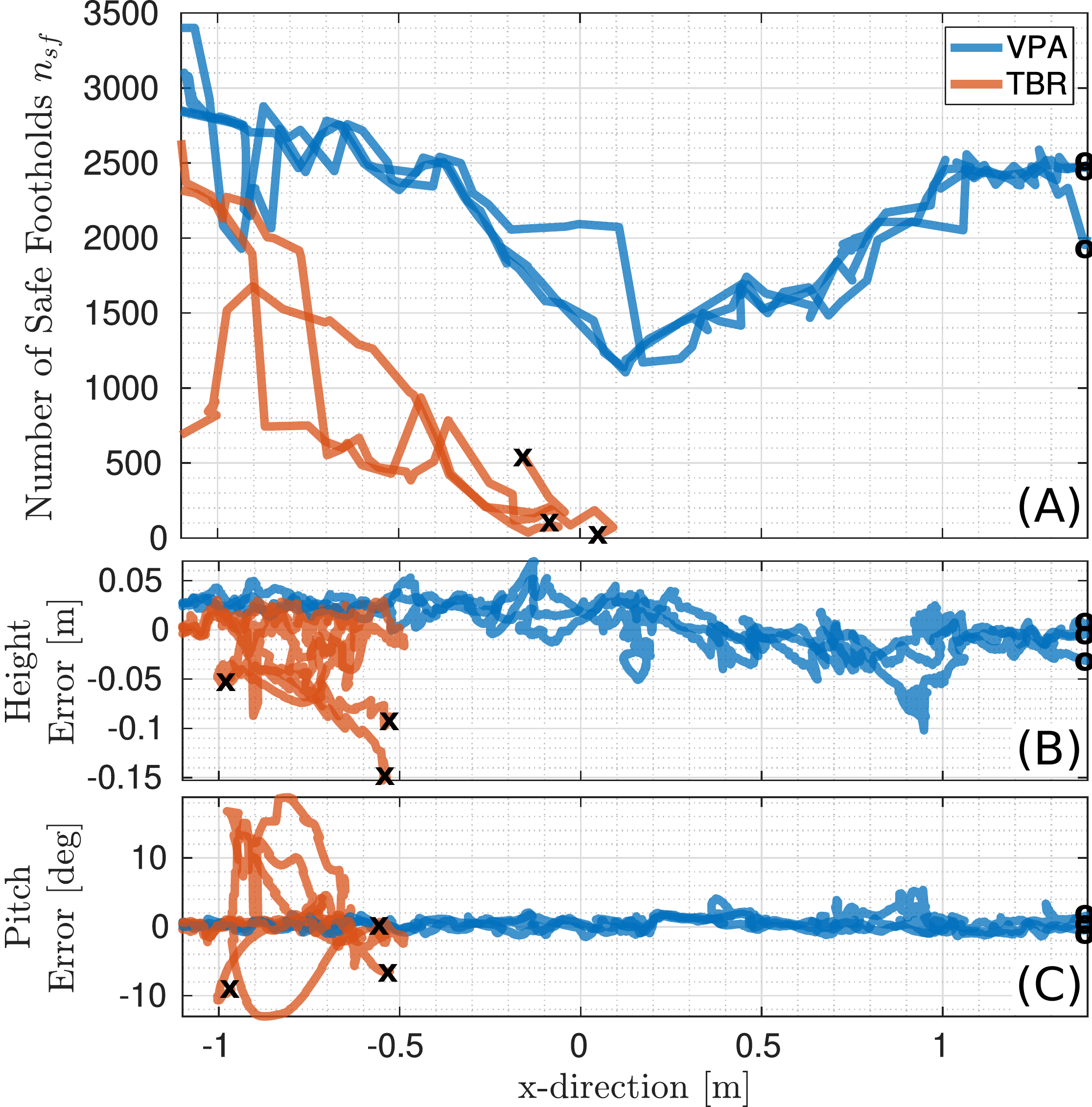}		
\caption
[The difference between the \acrshort{vpa} and the \acrshort{tbr}.]
{ 
The difference between the \acrshort{vpa} and the \acrshort{tbr} in six experiments (3 each).
(A)~The number of safe footholds corresponding to the robot pose.
(B,C)~The body height and pitch tracking errors, respectively. 
Circles~(\textbf{o})  and crosses (\textbf{x})  are successful and failed trials, respectively.
Unlike the~\acrshort{vpa}, the~\acrshort{tbr} failed to climb the stairs
because the~\acrshort{tbr} resulted in almost no safe footholds for the four legs to reach.
}
\label{fig_9}
\end{figure} 
We compare the \gls{vpa} with another vision-based pose adaptation strategy:
the \gls{tbr}~\cite{Villarreal2019}.
The \gls{tbr} generates pose references based on the footholds selected by the \gls{vfa}. 
The~\gls{tbr} fits a plane that passes through the given selected footholds, 
and sets the orientation of this plane as a body orientation reference to the robot.
The elevation reference of the \gls{tbr} is a constant distance
from the center of the approximated plane that passes through 
the selected footholds.
We chose the~\gls{tbr} instead of an optimization-based strategy 
since the latter does not provide references that are fast enough with respect to the~\gls{vpa}.

Using the stairs setup in~\fref{fig_1}(A), we conducted six experimental trials:
three with the~\gls{vpa} and three with the~\gls{tbr}.  
All trials were with the \gls{vfa}. 
In all trials, 
\acrshort{hyq} is commanded to crawl with a desired forward velocity of~\unit[0.1]{m/s}.
The results are reported in~\fref{fig_9} and~\vref{9}. 
Figure~\ref{fig_9}(A) shows the number of safe footholds corresponding to the robot pose
from the~\gls{vpa} and the~\gls{tbr}. 
The robot height and pitch tracking errors are shown in \fref{fig_9}(B,C), respectively.

As shown in~\vref{9}, \acrshort{hyq} failed to climb the stairs with the~\gls{tbr}, 
while it succeeded with the~\gls{vpa}. 
This is because, unlike the~\gls{vpa},
the~\gls{tbr} does not aim to 
put the robot in a pose that maximizes the chances of the legs to succeed in finding safe footholds. 
As shown in~\fref{fig_9}(A), the number of safe footholds from using the \gls{tbr} was below the ones from using the \gls{vpa}. 
During critical periods when the robot was around \unit[0]{m} in the x-direction,
the number of safe footholds from using the~\acrshort{tbr} almost reached zero.
The low number of safe footholds for the~\gls{tbr} compared to the~\gls{vpa}
is reflected in the tracking of the robot height and pitch as shown in 
{\fref{fig_9}(B,C)} 
where the tracking errors from the~\gls{tbr} were higher than the~\gls{vpa}.

The difference between the \gls{vpa} and the \gls{tbr} can be further explained in~\vref{9}.  
When the \gls{tbr} is used, the robot is adapting its pose \textit{given} the selected foothold. 
But, if the selected foothold is not reached, or if there is a high tracking error, 
the robot reaches a body pose that results in a smaller number of safe footholds. 
Thus, the feet end up colliding with the terrain and hence the robot falls. 
On the other hand, the~\gls{vpa} is able to put the robot 
in a pose that maximizes the number of safe footholds. 
As a result, the feet found alternative safe footholds to select from, 
which resulted in no collision, and succeeded in climbing the stairs.
The~\gls{vpa} optimizes for the number of safe footholds.
Thus, if there is a variation around the optimal pose (tracking error),
the~\gls{vfa} still finds more footholds to step on, which is not the case~with~the~\gls{tbr}.

\subsection{Comparing the \acrshort{vpa} with a Baseline (Simulation)}\label{tbr_result_sim}
\begin{figure}[t!]
\centering
\includegraphics[width=\columnwidth]{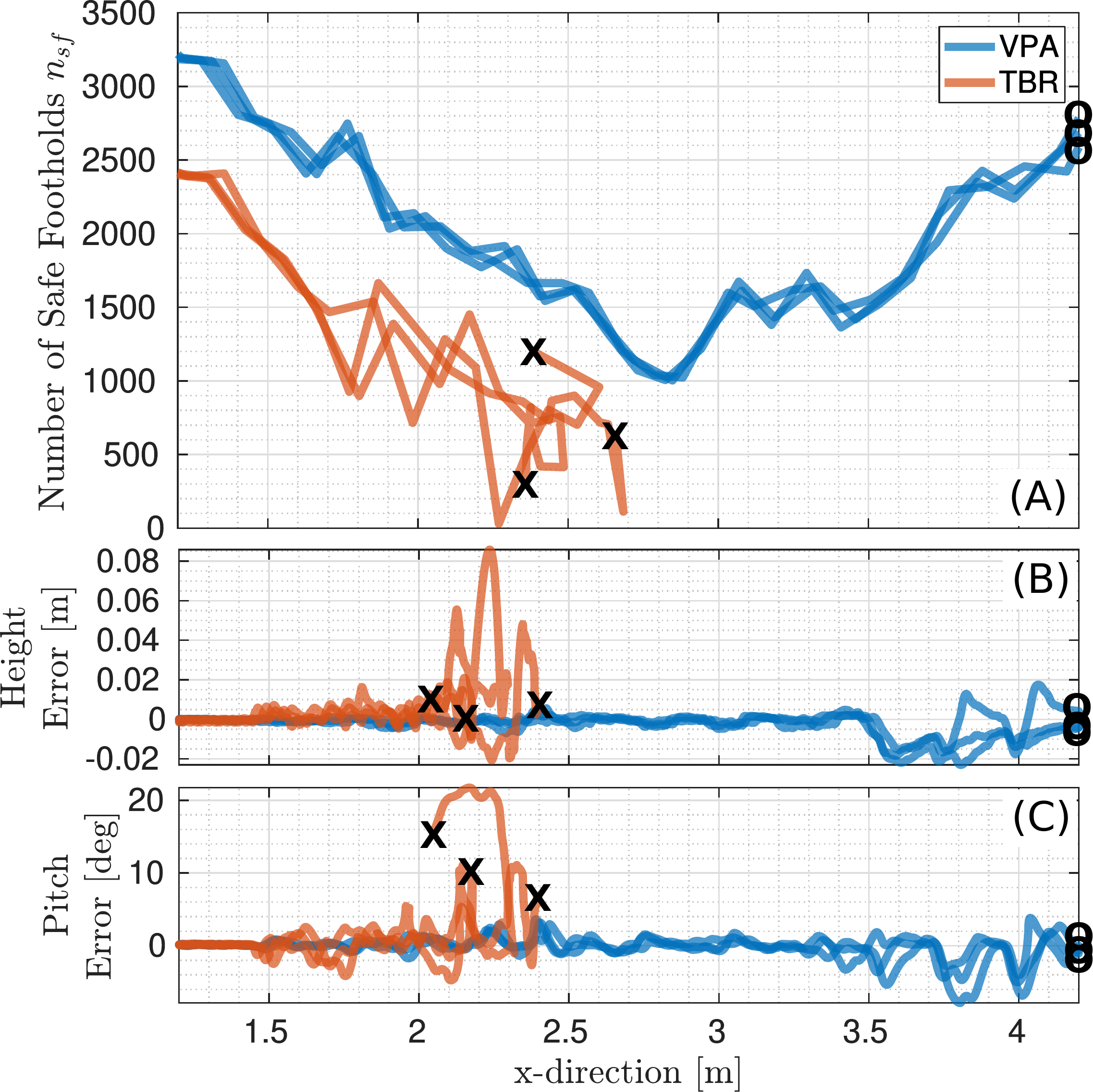}		
\caption[The difference between the \acrshort{vpa} and the \acrshort{tbr} in six simulations (3 each).]
{The difference between the \acrshort{vpa} and the \acrshort{tbr} in six simulations (3 each).
(A)~The number of safe footholds corresponding to the robot pose.
(B,C)~The body height and pitch tracking errors, respectively. 
Circles~(\textbf{o})  and crosses (\textbf{x})  are successful and failed trials, respectively.
Unlike the~\acrshort{vpa}, the~\acrshort{tbr} failed to climb the stairs
because the~\acrshort{tbr} resulted in almost no safe footholds for the four legs to reach.}
\label{fig_res_3_1}
\end{figure}
Similar to experiments, and using the stairs setup in~\fref{fig_4}, we compare  the~\acrshort{vpa} with  the~\acrshort{tbr}.
We conducted six simulations: three with the~\acrshort{vpa} and three with the~\acrshort{tbr}.  
All trials were with the \acrshort{vfa}. 
In all trials, \acrshort{hyq} is commanded to trot with a $0.2$\unit{m/s} desired forward velocity.
The results are reported in~\fref{fig_res_3_1} and~\vref{10}. 
Figure~\ref{fig_res_3_1}(A) shows the number of safe footholds corresponding to the robot pose from the~\acrshort{vpa}, and the~\acrshort{tbr}. 
The tracking errors of the robot height and pitch are shown in~\fref{fig_res_3_1}(B,C), respectively.
These trials show that~\acrshort{hyq} failed to climb the stairs using the~\acrshort{tbr}, while it succeeded using the~\acrshort{vpa}.

\subsection{Climbing Stairs with Gaps}\label{sec_sim_7}
\begin{figure}[t!]
\centering
\includegraphics[width=\columnwidth]{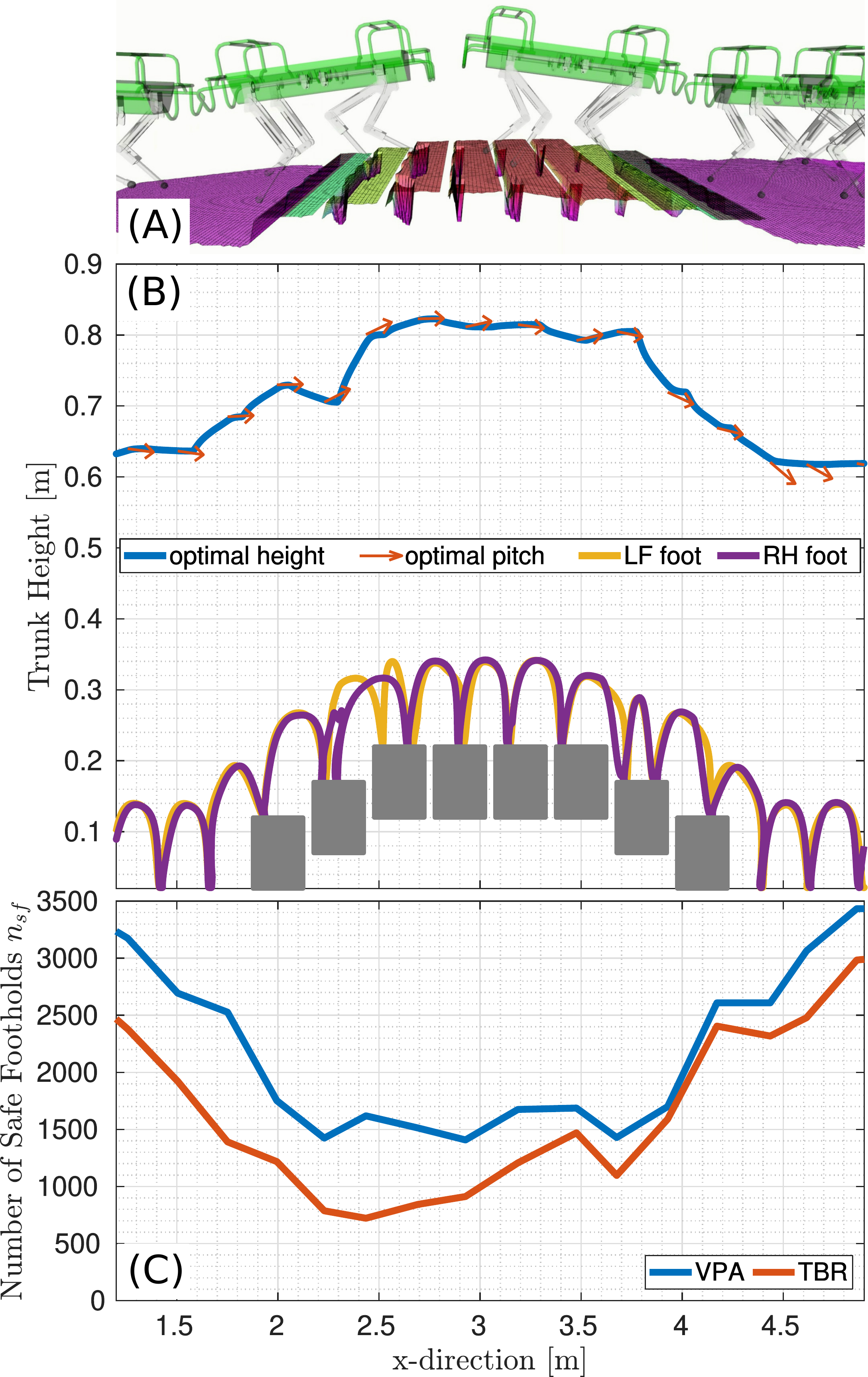}
\caption[\acrshort{hyq} climbing gapped stairs.]{\acrshort{hyq} climbing gapped stairs.
(A)~Screenshots of~\acrshort{hyq} climbing the setup.
(B)~The robot's height and pitch based on the~\acrshort{vpa}, and the corresponding feet trajectories of the~\acrshort{lf} and~\acrshort{rh} legs based on the~\acrshort{vfa}.
(C)~The number of safe footholds using the~\acrshort{vpa} and the~\acrshort{tbr}.}
\label{fig_5}
\end{figure} 
We show \acrshort{hyq}'s capabilities of climbing stairs with gaps using \acrshort{vital}, and we compare the~\acrshort{vpa} with the~\acrshort{tbr}.
In this scenario, \acrshort{hyq} is commanded to trot at~\unit[0.4]{m/s}.
Figure~\ref{fig_5}(A) shows overlayed screenshots of the simulation and the used setup.
Figure~\ref{fig_5}(B) shows the robot's height and pitch based on the~\acrshort{vpa}, 
and the corresponding feet trajectories of the~\acrshort{lf} leg and the~\acrshort{rh} leg based on the~\acrshort{vfa},
and Fig.~\ref{fig_5}(C) shows the number of safe footholds using the~\acrshort{vpa} and the~\acrshort{tbr}.
Because of~\acrshort{vital}, \acrshort{hyq} was able to climb the stairs with gaps 
while continuously adapting its pose and feet. 
Furthermore, the number of safe footholds from using the~\acrshort{tbr} is always lower than from using the~\acrshort{vpa},
which shows that indeed the~\acrshort{vpa} outperforms the~\acrshort{tbr}.
\vref{11} shows the output of this simulation using~\gls{vital}.

\subsection{Pose Optimization: Single vs. Receding Horizons}\label{sec_sim_5}
\begin{figure}[t!]
\centering
\includegraphics[width=\columnwidth]{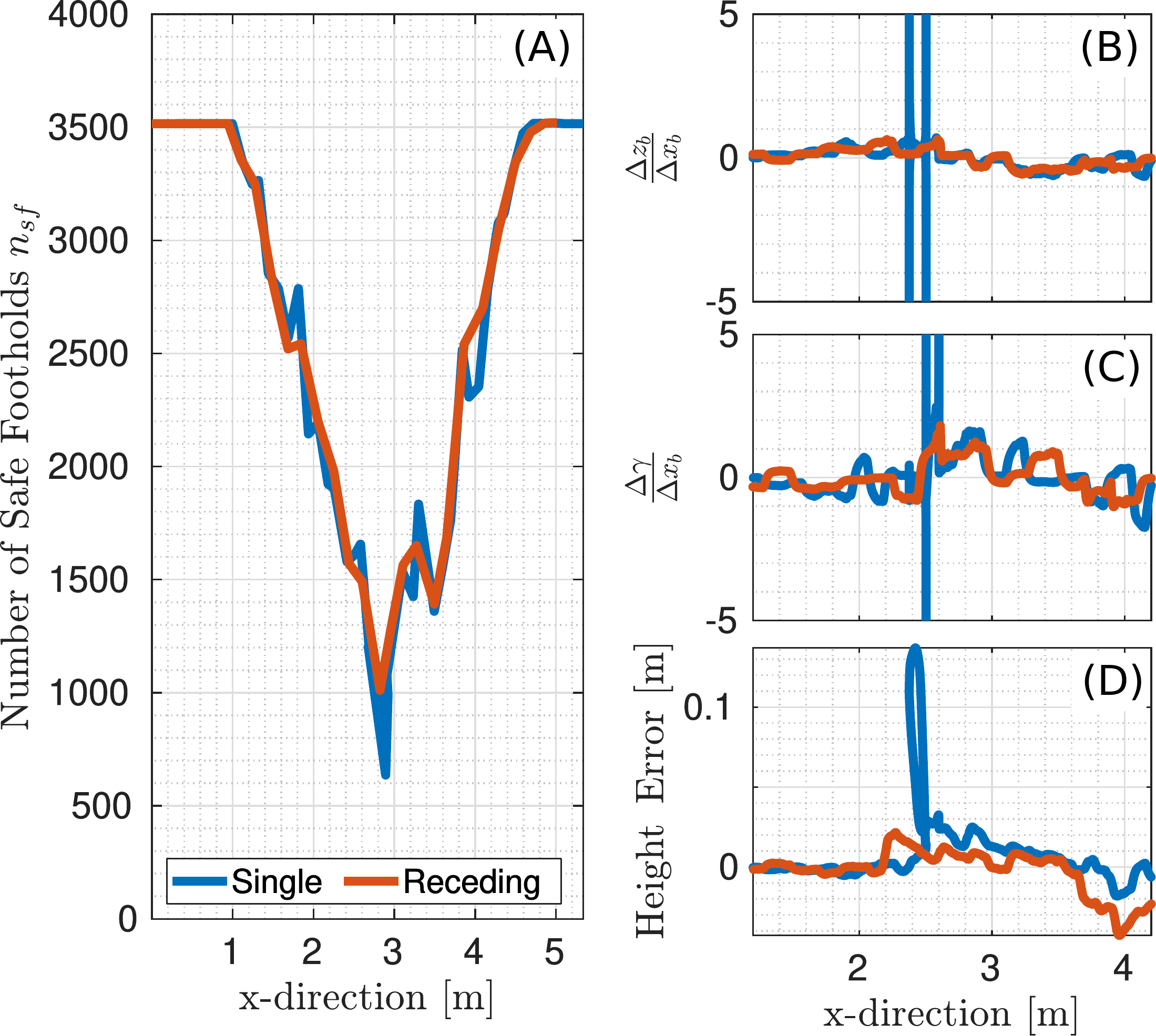}
\caption[Pose Optimization: Single vs. Receding Horizons.]
{Pose Optimization: Single vs. Receding Horizons.
(A)~The number of safe footholds.
(B)~Variation (numerical difference) of the robot's height. 
(C)~Variation (numerical difference) of the robot's pitch. 
(D)~The tracking error of the robot's height.}
\label{res_5_1}
\end{figure} 
To analyze the differences between the receding horizon and the single horizon 
in pose optimization, we use the stairs setup in~\fref{fig_4}
with a commanded forward velocity of~\unit[0.4]{m/s}, 
and report the outcome in~\fref{res_5_1} and~\vref{12}. 
The main advantage of using a receding horizon instead of a single horizon 
is that the pose optimization can consider future decisions. 
Thus, if the robot is trotting at higher velocities,
the pose optimizer can adapt the robot's pose before hand. 
This can result in a better adaptation strategy 
with less variations in the generated optimal pose.
Thus, we analyze the two approaches by taking a look at the variations in the body pose
\begin{equation}
\acute{z}_b = \frac{\Delta z_b}{\Delta x_b} 
~,~~~\text{and}~~~  
\acute{\gamma} = \frac{\Delta \gamma}{\Delta x_b}
\end{equation}
where $\acute{z}_b$ and $\acute{\gamma}$
are the numerical differences (variations) of 
the robot height $z_b$ and pitch $\gamma$ 
with respect to the robot forward position $x_b$, respectively.

Figure~\ref{res_5_1} reports the differences between the two cases.
The number of safe footholds is shown in~\fref{res_5_1}(A).
The variations in $\acute{z}_b$ and $\acute{\gamma}$ 
are shown in~\fref{res_5_1}(B,C), respectively. 
Finally, the tracking error of the robot's height is shown in~\fref{res_5_1}(D).
As shown in Figure~\ref{res_5_1} the receding horizon 
resulted in less variations in the body pose compared to 
the single horizon. 
This resulted in a smaller tracking error for the receding horizon
in the body height, which resulted in slightly larger number of safe footholds. 
All in all, the receding horizon reduces variations in the desired trajectories 
which improves the trajectory tracking response.

The differences between the receding and single horizon in the pose optimization
can also be noticed in~\vref{12}. 
In the case of a single horizon, 
the robot was struggling while climbing up the stairs but 
was able to recover and accomplish the task.
However, using the receding horizon, the robot was able 
to adapt its pose in time, 
and thus resulting in safer footholds
that allowed the robot to accomplish the task.

\begin{figure}[t!]
\centering
\includegraphics[width=\columnwidth]{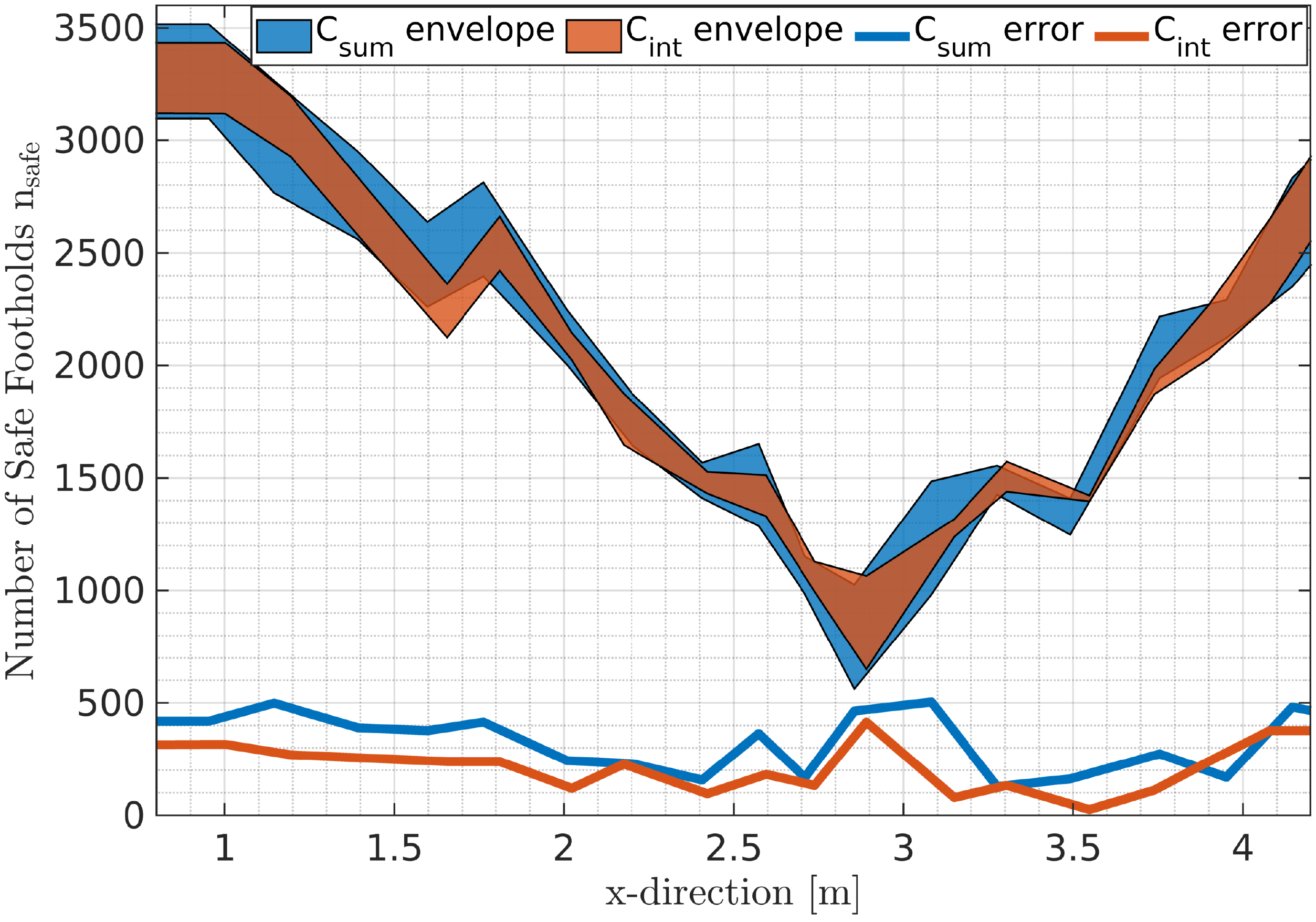}		
\caption[Pose Optimization: $\mathcal{C}_{\mathrm{sum}}$ vs. $\mathcal{C}_{\mathrm{int}}$.]
{Pose Optimization: $\mathcal{C}_{\mathrm{sum}}$ vs. $\mathcal{C}_{\mathrm{int}}$.
The shaded areas are the envelopes of the number of safe footholds. 
The lines are the thicknesses (errors) between these envelopes.}
\label{res_6_1}
\end{figure} 

\subsection{Pose Optimization: $\mathcal{C}_{\mathrm{sum}}$ vs. $\mathcal{C}_{\mathrm{int}}$}
\label{sec_sim_6}

To analyze the differences between $\mathcal{C}_{\mathrm{sum}}$ and $\mathcal{C}_{\mathrm{int}}$
in the pose optimization, 
we use the stairs setup in~\fref{fig_4} with a commanded forward velocity of~\unit[0.4]{m/s}, 
and report the outcome in~\fref{res_6_1}.
The main advantage of using $\mathcal{C}_{\mathrm{int}}$
over $\mathcal{C}_{\mathrm{sum}}$ is that $\mathcal{C}_{\mathrm{int}}$
will result in a pose that does not just maximize the number of safe footholds for all of the legs, 
but also ensures that the 
number of safe footholds of the poses around the optimal pose is still high.
To compare the two cost functions, 
we take a look at the number of safe footholds. 
In particular, we evaluate the number of safe footholds 
corresponding to the optimal pose, and the poses around it with a margin of $m=$~\unit[0.025]{m}.
Thus, in~\fref{res_6_1}, we plot the envelope (shaded area) between 
$\rbfn(u^* + m)$ and $\rbfn(u^* - m)$ for both cases, and the thickness between these envelopes 
which we refer to as error 
\begin{equation}
\textbf{error}~=\vert \rbfn(u^* + m)-\rbfn(u^* - m)\vert .
\end{equation}

As shown in~\fref{res_6_1}
the envelope of the number of safe footholds 
resulting from $\mathcal{C}_{\mathrm{int}}$
is almost always encapsulated by $\mathcal{C}_{\mathrm{sum}}$.
The thickness (error) of
the number of safe footholds resulting from using $\mathcal{C}_{\mathrm{int}}$ 
is always smaller than $\mathcal{C}_{\mathrm{sum}}$.
This means that any variation of $m$ in the optimal pose
will be less critical if  $\mathcal{C}_{\mathrm{int}}$ is used compared to 
$\mathcal{C}_{\mathrm{sum}}$.

\begin{figure}[t!]
\centering
\includegraphics[width=\columnwidth]{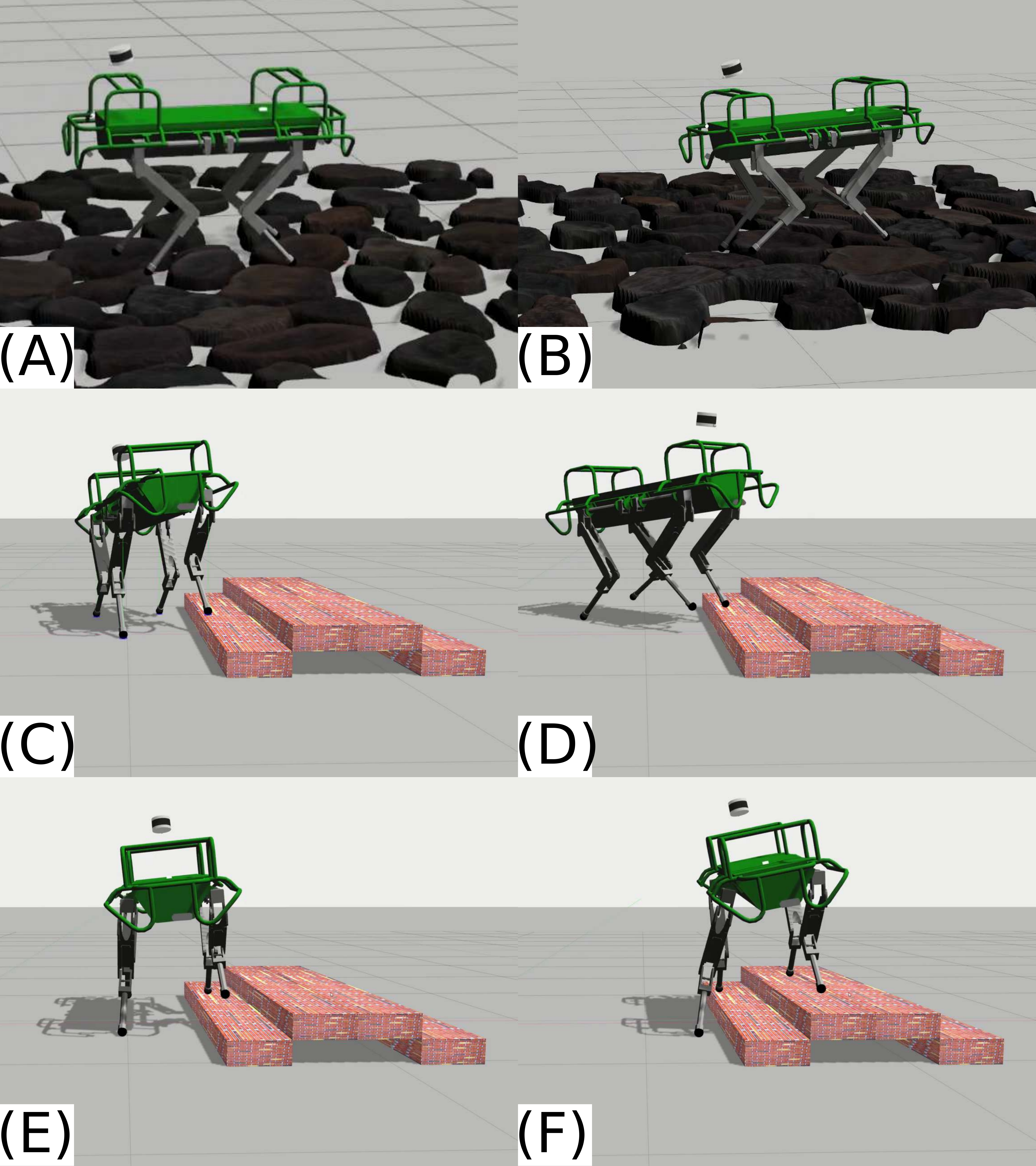}		
\caption[\acrshort{hyq} traversing rough terrain and climbing stairs sideways using~\gls{vital}.]
{\acrshort{hyq} traversing rough terrain and climbing stairs sideways.
(A,B)~\acrshort{hyq} traversing rough terrain with and without~\gls{vital}, respectively.
(C,D)~\acrshort{hyq} climbing stairs while yawing (commanding the yaw rate) using~\gls{vital}.
(E,F)~\acrshort{hyq} climbing stairs sideways using~\gls{vital}.}
\label{res_15}
\end{figure} 

\subsection{Locomotion over Rough Terrain}
We evaluate the performance of~\acrshort{hyq} in traversing 
rough terrain as shown in~\fref{res_15}(A,B) and in~\vref{13}. 
We conducted two simulations:
one with~\gls{vital} and thus with exteroceptive and proprioceptive reactions~(\fref{res_15}(A)),
and another without~\gls{vital} and thus only with proprioceptive reactions~(\fref{res_15}(B)). 
\acrshort{hyq} was commanded to traverse the rough terrain with a forward velocity of~\unit[0.2]{m/s}.
No hyper parameters re-tuning, or \glspl{cnn} re-training were needed.

As shown in~\vref{13},~\acrshort{hyq} was able to successfully traverse the terrain in both cases. 
With~\gls{vital},~\acrshort{hyq} collided less with the terrain and continuously adapted its footholds over the small cobblestones. 
Without~\gls{vital},~\acrshort{hyq} traversed the rough terrain, yet, with significantly more effort. 
Additionally, without~\gls{vital},~\acrshort{hyq} continuously collided with the terrain, and in some incidents, the feet got stuck. 
For this reason, we had to re-tune the gait parameters, and increase the step height to reduce these incidents. 
The robot's feet also kept slipping since the feet were always close to edges and corners.

\subsection{Climbing Stairs with Different Commands}
Instead of commanding only forward velocities as in the previous sections, 
we command~\acrshort{hyq} to climb the stairs with~\gls{vital}
while yawing (commanding the yaw rate) as shown in~\vref{14} and~\fref{res_15}(C,D), and 
to climb stairs laterally as shown in~\vref{15} and~\fref{res_15}(E,F). 
Climbing stairs sideways is more challenging than facing the stairs 
since the range of motion of the robot's roll orientation is more restricted versus the pitch orientation. 
That said, because of~\gls{vital}, \acrshort{hyq} was still able to climb these stairs in both cases as shown in~\vref{14} and~\vref{15}.

\vspace{10pt}
\section{Conclusion}\label{sec_conclusion_vital}
We presented \gls{vital} which is an online vision-based locomotion planning strategy.
\gls{vital} consists of the~\gls{vpa} for pose adaptation, 
and the~\gls{vfa} for foothold selection. 
The~\gls{vpa} introduces a different paradigm 
to current state-of-the-art pose adaptation strategies. 
The~\gls{vpa} finds body poses that maximize the chances 
of the legs to succeed in reaching safe footholds. 
This notion of success emerges from the robot's skills. 
These skills are encapsulated in the \gls{fec}
that include (but are not limited to)
the terrain roughness, kinematic feasibility, leg collision, and foot trajectory collision. 
The~\gls{vfa} is a foothold selection algorithm
that continuously adapts the robot's trajectory based on the~\criteria.
The~\gls{vfa} algorithm of this work extends 
our previous work in~\cite{Esteban2020, Villarreal2019} as well as the state of the art~\cite{Fankhauser2018,Jenelten2020}. 
Since the computation of the~\criteria is usually expensive, 
we rely on approximating these criteria with~\glspl{cnn}.

The robot's skills and the notion of success provided 
by the \gls{fec} allowed the~\gls{vpa}
to generate body poses that maximize the chances of success in reaching safe footholds. 
This resulted in body poses that are aware of the terrain
and aware of what the robot and its legs can do. 
For that reason, 
the~\gls{vpa} was able to generate body poses that give a better chance
for the~\gls{vfa} to select safe footholds. 
As a result, because of~\gls{vital}, 
\acrshort{hyq} and \acrshort{hyqreal} were able to traverse
multiple terrains with various forward velocities and different gaits
without colliding or reaching workspace limits. 
The terrains included stairs, gaps, and rough terrains, 
and the commanded velocities varied from \unit[0.2]{m/s} to~\unit[0.75]{m/s}.
The~\gls{vpa} outperformed other strategies for pose adaptation. 
We compared~\gls{vpa}
with the~\gls{tbr} which is another vision based pose adaptation strategy,
and showed that indeed the~\gls{vpa} puts the robot
in a pose that provides the feet with higher number of safe footholds. 
Because of this, 
the~\gls{vpa} made our robots succeed in various scenarios where the~\gls{tbr} failed.

\section{Limitations and Future Work}
One issue that we faced during experiment was in tracking the motion of the robot, especially for~\acrshort{hyqreal}. 
We were using a~\gls{wbc} for motion tracking.  
We believe that the motion tracking and our strategy can be improved by using a~\gls{mpc} alongside the~\gls{wbc}.
Similarly, instead of using a model-based controller (\gls{mpc} or \gls{wbc}), 
we hypothesize that an~\gls{drl}-based controller can also improve the robustness and reliability of the overall 
robot behavior. 

As explained in~\sref{sec_sys_over}, 
one other key limitation was regarding the perception system. 
State estimation introduced a significant drift that caused a major noise and drift in the terrain map. 
Albeit not being a limitation to the suggested approach, 
we plan on improving the state estimation and perception system of~\acrshort{hyq} and~\acrshort{hyqreal}
to allow us to test~\gls{vital} in the wild.

The pose optimization problem of the~\gls{vpa} does not reason about the robot's dynamics. 
This did not prevent~\acrshort{hyq} and~\acrshort{hyqreal}
from achieving dynamic locomotion while traversing challenging terrains at high speeds. 
However, we believe that 
incorporating the robot's dynamics into~\gls{vital} may result in a better overall performance. 
That said, we believe that in the future, the~\gls{vpa} should also reason about the robot's dynamics. 
For instance, 
one can augment the \criteria with another criterion that ensures that the selected footholds are dynamically feasible by the robot.

Additionally, in the future,
we plan to extend the~\gls{vpa} of~\gls{vital} to not only send pose references, 
but also reason about the robot's body twist. 
We also plan to augment the robot skills to not only consider foothold evaluation criteria, 
but also skills that are tailored to the robot pose. 
Finally, in this work, \gls{vital} considered heightmaps which are 2.5D maps. 
In the future, we plan to consider full 3D maps that will enable~\gls{vital} 
to reason about navigating in confined space (inspired by \cite{Buchanan2020b}).

\appendix
\subsection{CNN approximation in the VFA and the VPA}\label{cnn_approx}
In the~\gls{vfa}, the foothold evaluation stage is approximated with a \acrshort{cnn}~\cite{Lecun1989} as explained in~\remref{remark_cnn_exact}.
The \acrshort{cnn} approximates the mapping between~$\heurtuple$ and~$\optimal$.
The heightmap~$\hmapvfa$~in~$\heurtuple$ passes through three convolutional layers 
with~$5 \times 5$ kernels, $2 \times 2$ padding, Leaky ReLU activation~\cite{Maas2013}, and $2 \times 2$ max-pooling operation.
The resulted one-dimensional feature vector is concatenated with 
the rest of the variables in the tuple~$\heurtuple$,
namely, $\hheight, \bodyvel, \gaitparams,$ and $\nominal$. 
This new vector passes through two fully-connected layers with Leaky ReLU and softmax activations.
The parameters of the \acrshort{cnn} are optimized to minimize the cross-entropy loss~\cite{Bishop2006} of classifying a candidate foothold location as optimal~$\optimal$.

In the~\gls{vpa}, the pose evaluation and the function approximation is approximated with a \acrshort{cnn} as explained in~\remref{remark_cnn_exact_vpa}.
The \acrshort{cnn} infers the weights~$w$ of~$\rbfn$ given~$\heurtuplevpa$ (the mapping between~$\heurtuplevpa$ and~$w$).
The heightmap $H_{\mathrm{vpa}}\in\Rnum^{33\times33}$ passes through three convolutional layers 
with $5 \times 5$ kernels, $2 \times 2$ padding, Leaky ReLU activation, and $2 \times 2$ max-pooling operation.
The body velocities~$\bodyvel$ pass through a fully-connected layer with Leaky ReLU activation 
that is then concatenated with the one-dimensional feature vector obtained from the heightmap.
This new vector passes through two fully-connected layers with Leaky ReLU and linear activations.
The parameters of this \acrshort{cnn} are optimized 
to minimize the mean squared error loss
between the number of safe footholds~\acrshort{nsf} predicted by $\rbfn(z_{h},w)$ 
and $\rbfn(z_{h},\hat{w})$ where $\hat{w}$ 
are the function parameters approximated by the~\acrshort{cnn}.

For both \glspl{cnn}, we used the Adam optimizer~\cite{Kingma2015} with a learning rate of $0.001$,
and  we used a validation-based early-stopping using a $9$-to-$1$~proportion to reduce overfitting.
The datasets required for training this \glspl{cnn} are collected 
by running simulated terrain scenarios that consist of bars, gaps, stairs, and rocks. 
In this work, we considered a $33\times33$ heightmap with a resolution of \unit[0.02]{m} 
($H_{\mathrm{vfa}}, H_{\mathrm{vpa}}\in\Rnum^{33\times33}$).

\begin{figure}[t!]
\centering
\includegraphics[width=\columnwidth]{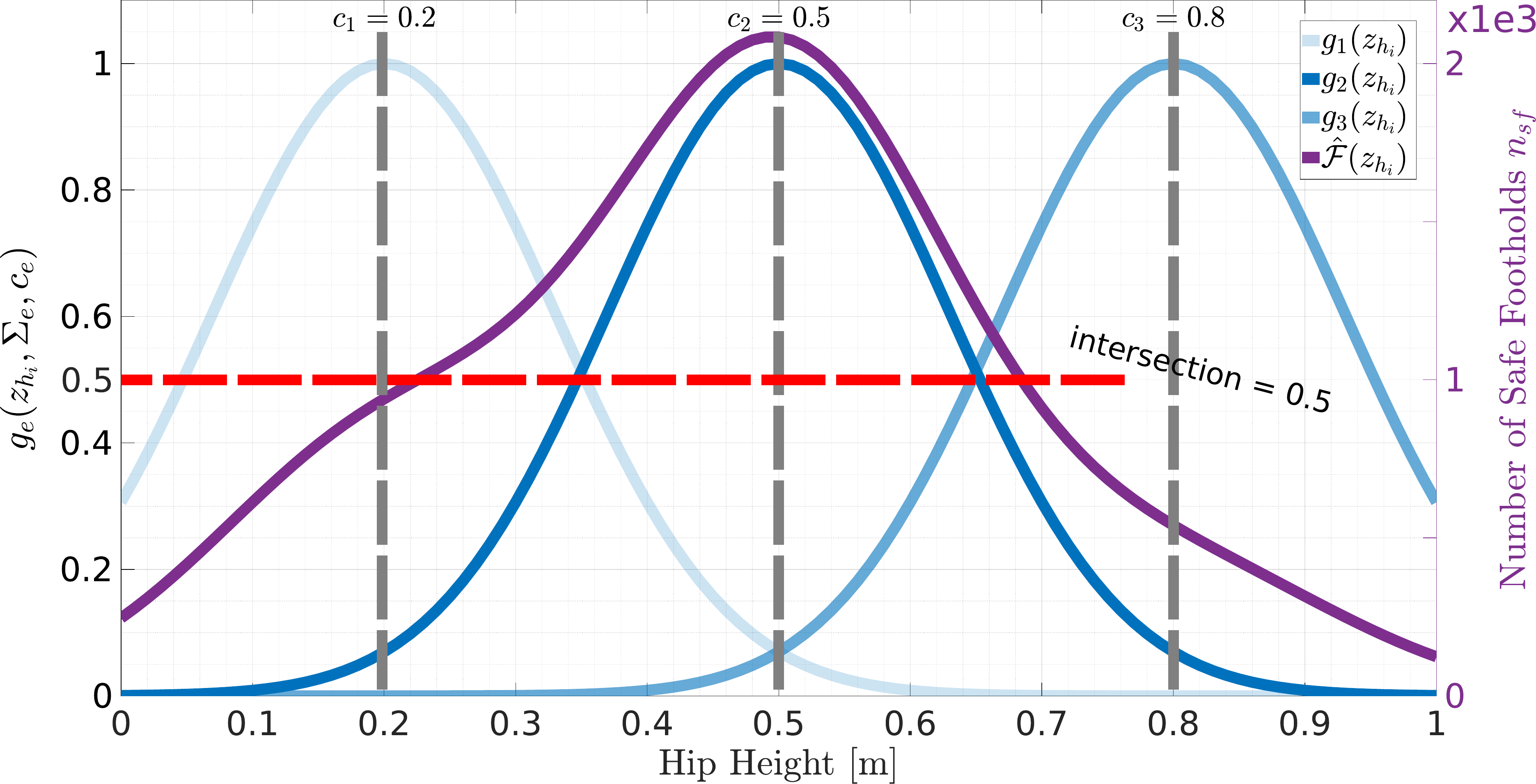}		
\caption{An illustration of the Function Approximation of the \acrshort{vpa}.}
\label{fig_func_approx}
\end{figure} 

\subsection{Details on the Function Approximation of the \acrshort{vpa}}\label{app_func_approx}
As explained in~\sref{sec_func_approx}, the function~$\rbfn(z_{h_i}, w)$ 
\begin{equation}
\rbfn(z_{h_i}, w) = \sum_{e=1}^{E} w_e \cdot g(z_{h_i}, \Sigma_e, c_e)
\end{equation}
is defined as the weighted sum of Gaussian basis functions
\begin{equation}
g(z_{h_i}, \Sigma_e, c_e) = \text{exp}(-0.5 (z_{h_i}-c_e)^T \Sigma_e^{-1} (z_{h_i} -c_e)).
\end{equation}
The parameters $\Sigma_e$ and $c_e$ are the widths and centers of the Gaussian function~$g_e$~(see Section~3.1 in~\cite{Stulp2015}).
In the literature,~$c_e$ 
is usually referred to as the mean or the expected value, and $\Sigma_e$ as the standard deviation.
The regression algorithm should predict the weights~$w_e$, and the parameters~$\Sigma_e$ and~$c_e$.
To reduce the dimensionality of the problem, as explained in~\sref{sec_func_approx}, and in~Section~4.1 in~\cite{Stulp2015},
we decided to fix the values of the parameters of the Gaussian functions~$\Sigma_e$ and~$c_e$.
In detail, 
the centers~$c_e$ are spaced equidistantly within the bounds of the hip heights~$\elementheights$,
and the widths are determined by the value at which the Gaussian functions intersect.
That way, the regression algorithm only outputs the weights~$w_e$.
Figure~\ref{fig_func_approx} shows an example of the function approximation. 
In this example, the bounds of the  hip heights~$\elementheights$ are \unit[0.2]{m} and \unit[0.8]{m}.
Assuming a number of basis functions~$E=3$, the centers~$c_e$ are then chosen to be equidistant within the bounds, 
and thus, the centers~$c_e$ are \unit[0.2]{m}, \unit[0.3]{m} and \unit[0.8]{m}. 
By choosing the Gaussian functions to intersect at 0.5, the widths~$\Sigma_e$ are~0.13.

\subsection{Representing the Hip Heights in terms of the Body Pose}\label{po_appendix}
To represent the hip heights in terms of the body pose, 
we first write the forward kinematics of the robot's hips
\begin{equation}p_{h_i}^W = p_{b}^W ~+~ R_b^W p_{h_i}^b\label{hh_respresentation_1}\end{equation}
where $p_{h_i}^W\in\Rnum^3$~is the position of the hip of the $i$th leg in the world frame, 
$p_{b}^W\in\Rnum^3$~is the position of the robot's base in the world frame,
$R_b^W\in SO(3)$~is the rotation matrix mapping vectors from the base frame to the world frame,
and $p_{h_i}^b\in\Rnum^3$~is the position of the hip of the $i$th leg in the base frame.
The rotation matrix $R_b^W$ is a representation of the Euler angles of the robot's base
with sequence of roll~$\beta$, pitch~$\gamma$, and yaw~$\psi$ (Cardan angles)~\cite{Diebel2006}.
The variable $p_{h_i}^b$ is obtained from the CAD of the robot. 
Expanding~\eref{hh_respresentation_1} yields
\begin{eqnarray}
\begin{bmatrix} x_{h_i}^W \\ y_{h_i}^W \\ z_{h_i}^W \end{bmatrix}
&\hspace{-5pt}=\hspace{-5pt}& 
\begin{bmatrix} x_{b}^W \\ y_{b}^W \\ z_{b}^W \end{bmatrix} + R_b^w (\beta, \gamma, \psi) \begin{bmatrix} x_{h_i}^b \\ y_{h_i}^b \\ z_{h_i}^b \end{bmatrix}\\
&\hspace{-5pt}=\hspace{-5pt}& 
\begin{bmatrix} x_{b}^W \\ y_{b}^W \\ z_{b}^W \end{bmatrix} 
+ \begin{bmatrix} \cdots  & \cdots & \cdots \\ \cdots  & \cdots & \cdots \\ -s\gamma & c\gamma~s\beta &  c\gamma~c\beta\end{bmatrix}
\begin{bmatrix} x_{h_i}^b \\ y_{h_i}^b \\ z_{h_i}^b \end{bmatrix} 
\label{hh_respresentation_2}
\end{eqnarray}
where $s$ and $c$ are sine and cosine of the angles, respectively. 
Since we are interested only in the hip heights, 
the z-component (third row) of~\eref{hh_respresentation_2} yields
\begin{equation}
z_{h_i}^W = z_{b}^W - x_{h_i}^b s\gamma + y_{h_i}^b c\gamma s\beta + z_{h_i}^b c\gamma c\beta.
\end{equation}

\subsection{Details on Using the Sum of Squared Integrals as a Cost Function in the Pose Optimization Problem of the~\gls{vpa}}\label{cint_appendix}
\begin{figure}[t!]
\centering
\includegraphics[width=\columnwidth]{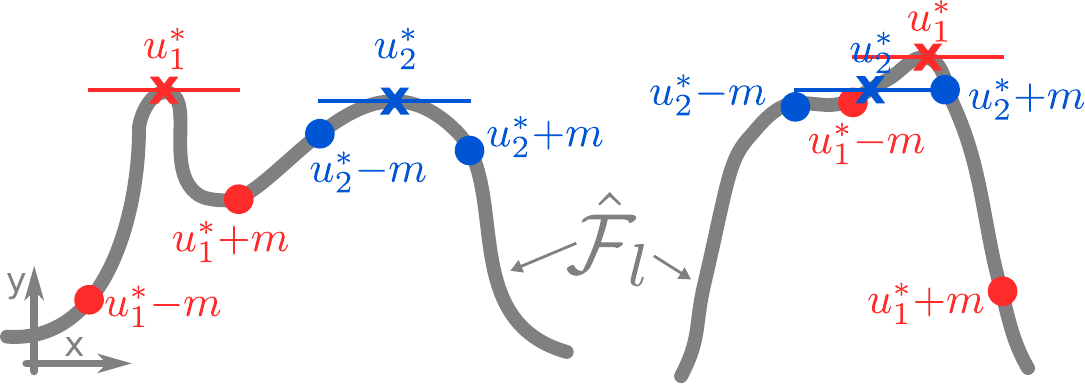}		
\caption[Using the sum of squared integrals as a cost function in the pose optimization of the~\gls{vpa}]
{Using the sum of squared integrals as a cost function in the pose optimization of the~\gls{vpa}.
The two curves represent~$\rbfn_l$. The x-axis is the hip height and the y-axis is~\acrshort{nsf}.
The figure shows two optimal poses: $u_1^*$~which is from 
using~$\mathcal{C}_{\mathrm{sum}}$ or~$\mathcal{C}_{\mathrm{prod}}$,
and $u_2^*$~which is from using~$\mathcal{C}_{\mathrm{int}}$.}
\label{fig_po}
\end{figure}
Using~$\mathcal{C}_{\mathrm{int}}$ as a cost function 
can be motivated by taking~\fref{fig_po} as an example. 
In this figure, 
there are two curves that represent~$\rbfn_l$
where the horizontal axis is the hip height and the vertical axis represent~\acrshort{nsf}.
The figure shows two optimal poses where $u_1^*$~is the optimal pose
using the cost functions~$\mathcal{C}_{\mathrm{sum}}$  or $\mathcal{C}_{\mathrm{prod}}$ that only maximize for~$\rbfn_l$,
and $u_2^*$~is the optimal pose using the cost function~$\mathcal{C}_{\mathrm{int}}$.
As shown in the figure,
if $\mathcal{C}_{\mathrm{sum}}$ or $\mathcal{C}_{\mathrm{prod}}$
is used, the optimal pose will be $u_1^*$
which is indeed the one that results in the maximum~$\rbfn_l$.
However, if there is a tracking error of $m$ 
(thus the robot reaches $u_1^* \pm m$), 
the robot might end up in the pose $u_1^* \pm m$
that results in a small number of safe footholds.
Using 
$\mathcal{C}_{\mathrm{int}}$ will take into account the safe footholds within a margin $m$.
This might result in a pose that does not yield the maximum number of safe footholds, 
but it will result in a safer foothold in case the robot pose has any tracking errors. 
Note that using
the sum of squared integrals~$\mathcal{C}_{\mathrm{int}}$
as a cost function
is similar to smoothing~$\rbfn_l$ with respect to the hip height 
(a moving average smoothing).
Using a smoothing function~$\mathcal{C}_{s}$ may take the form
\begin{equation}
\mathcal{C}_{s} = \sum_{l=1}^{N_l=4} \Vert  
\frac{1}{2\epsilon}  
\sum_{i=-\epsilon}^{\epsilon}  \rbfn_l (z_{h_l}+i) \
\Vert^2_Q.
\end{equation}

\subsection{Defining the Receding Horizon}\label{receding_def}
In the receding horizon there is a tuple~$\heurtuplevpaR$ for every $j$th horizon
that is defined~as
\begin{equation}
\heurtuplevpaR = (\hmapvpaR,\bodyvel, \gaitparams)
\label{vpa_tuble_receding}
\end{equation}
hence sharing the same body twist~$\bodyvel$ and gait parameters~$\gaitparams$ but a different heightmap~$\hmapvpaR$.
For every leg, a heightmap of horizon~$j+1$ is overlapping with  the previous horizon's~$j$ heightmap.
This overlap has a magnitude of~$\Delta h$ taking the same direction as the body velocity~$\dot{x}_b$.
Without loss of generality, we chose the magnitude of the overlap to be half of the diagonal size of the heightmap in this work. 
To sum up, we first gather~$\heurtuplevpaR$ that share the same~$\bodyvel$ and~$\gaitparams$, but a different~$\hmapvpaR$.
Then, we evaluate~$\heurtuplevpaR$ and approximate the output using the function approximation 
yielding~$\rbfn_{j}$ that is sent to the optimizer for all of the legs.

\subsection{Miscellaneous Settings}\label{misc_details}
In this work, all simulations were conducted on an Intel Core i7 quad-core CPU, 
and all experiments were running on an onboard Intel Core i7 quad-core CPU where state estimation, mapping, and controls were running. 
The~\gls{rcf} (including the~\gls{wbc}) runs at \unit[250]{Hz}, 
the low-level controller runs at \unit[1000]{Hz},
the state estimator runs at \unit[333]{Hz}, 
and the mapping algorithm runs at \unit[20]{Hz}.
The~\gls{vpa} and the~\gls{vfa} run asynchronously at the
maximum possible update rate.

\acrshort{vital} is implemented in Python. 
The~\acrshortpl{cnn} are implemented in PyTorch~\cite{Adam2019}.
As explained in~\appref{cnn_approx}, in this work, we considered a $33\times33$ heightmap with a resolution of \unit[0.02]{m} 
($H_{\mathrm{vfa}}, H_{\mathrm{vpa}}\in\Rnum^{33\times33}$).
The finite set~$\finitesetheights$
consisted of a hip height range between~\unit[0.2]{m} and~\unit[0.8]{m}
with a resolution of~\unit[0.02]{m} yielding~$\hipheightsamples=31$ samples. 
The number of radial basis functions used in the function approximation was $E=30$.
The pose optimization problem is 
solved with a 
trust-region interior point method~\cite{Wright1996,Byrd1999} which is a non-linear optimization problem solver that we solved using
SciPy~\cite{Virtanen2020}.
The bounds of the pose optimization problem~$u_{\min}$
and~$u_{\max}$ are~$[\unit[0.2]{m},\unit[-0.35]{rad},\unit[-0.35]{rad}]$, and~$[\unit[0.8]{m},\unit[0.35]{rad},\unit[0.35]{rad}]$, respectively.
We used a receding horizon of~$N_h=2$ with a map overlap of half the size of the heightmap.
For a heightmap of a size of~$33\times33$ and a resolution of~\unit[0.02]{m}, the map overlap~$\Delta h$ is~\unit[0.33]{m}.
We used Gazebo~\cite{Koenig2004} for the simulations, and ROS for communication.

\subsection{Estimation Accuracy}\label{est_acc}
We compare the estimation accuracy of the~\gls{vfa}
by comparing the output of the foothold evaluation stage (explained in \sref{vfaa}) given the same input tuple $\heurtuple$.
That is to say, we compare the estimation accuracy of the~\gls{vfa} by comparing $\ghat$ versus $\g$ (see \remref{remark_cnn_exact}).
To do so, once trained, we generated a dataset of \unit[4401]{samples} from randomly sampled heightmaps for every leg. 
This analysis was done on~\acrshort{hyq}.

As explained in~\sref{vfaa}, 
from all of the safe candidate footholds in~$\fecout$, 
the~\gls{vfa} chooses the optimal foothold to be the one closest to the nominal foothold. 
Thus, to fairly analyse the estimation accuracy of the~\gls{vfa}, we present three main measures: 
\textit{perfect match} being the amount of samples where $\ghat$ outputted the exact value of $\g$,
\textit{safe footholds}, being the amount of samples where  $\ghat$ did not output the exact value of $\g$, but rather a foothold that is safe but not closest to the nominal foothold, 
and \textit{mean distance}, being the average distance of the estimated optimal foothold from $\ghat$ relative to the exact foothold from $\g$. 
These measures are presented as the mean of all legs. 

Based on that, 
the perfect match measure is~$74.0\%$. 
Thus, $74\%$ of $\ghat$ perfectly matched $\g$. 
The safe footholds measure is~$93.7\%$. 
Thus, $93.7\%$ of $\ghat$ were deemed safe.
Finally, the mean distance of the estimated optimal foothold from $\ghat$ relative to the exact foothold from $\g$ is~\unit[0.02]{m}.
This means that, on average, $\ghat$ yielded optimal footholds that are~\unit[0.02]{m} far from the optimal foothold from $\g$.
Note that, the radius of~\acrshort{hyq}'s foot, and the resolution of the heightmap is also~\unit[0.02]{m}, 
which means that the average distance measure is still acceptable especially since we account for this value in the uncertainty margin as explained 
in \remref{remark_uncertainty_margin}.

Similar to the~\gls{vfa}, we compare the accuracy of~\gls{vpa}
by comparing  the output of the pose evaluation stage (explained in \sref{sec_vpa_pipeline}) given the same input tuple $\heurtuple$.
That is to say, we compare the estimation accuracy of the~\gls{vpa} by comparing $\finitesetfeasibles$ versus $\rbfn$ 
(see \remref{rem3} and \remref{remark_cnn_exact_vpa}).
To do so, we ran one simulation using the stairs setup shown in~\fref{fig_4} on~\acrshort{hyq}, and gathered the input tuple $\heurtuple$. 
Then, we ran the~\gls{vpa} offline, once with the exact evaluation (yielding~$\finitesetfeasibles$) 
and once with the approximate one (yielding~$\rbfn$).

Based on this simulation run, 
the mean values of the exact and the approximate evaluations are 
$\mathrm{mean}(\finitesetfeasibles) = 1370$ and
$\mathrm{mean}(\rbfn) = 1322$, respectively. 
This yields an estimation accuracy 
$\mathrm{mean}(\rbfn)/\mathrm{mean}(\finitesetfeasibles)$ 
of~$96.5\%$.

\subsection{Computational Analysis}\label{comp_anal}
To analyze the computational time of the~\gls{vfa} and the~\gls{vpa}, 
we ran one simulation using the stairs setup shown in~\fref{fig_4} on~\acrshort{hyq} 
to gather the input tuples of the~\gls{vfa} and the~\gls{vpa}, $\heurtuple$ and $\heurtuplevpa$, respectively. 
Then, we ran both algorithms offline, once with the exact evaluation and once using the~\glspl{cnn}, 
and collected the time it took to run both algorithms (all stages included).
The mean and standard deviation of the time taken to compute the exact and the~\gls{cnn}-approximated \gls{vfa} (per leg) algorithms are
$\unit[7.5]{ms} \pm \unit[1]{ms}$, and
$\unit[3.5]{ms} \pm \unit[1]{ms}$, respectively. 
The mean and standard deviation of the time taken to compute the exact and the~\gls{cnn}-approximated \gls{vpa} algorithms are
$\unit[720]{ms} \pm  \unit[68]{ms}$, and
$\unit[180]{ms} \pm \unit[60]{ms}$, respectively. 
Hence, the~\gls{vfa} and the~\gls{vpa} can run at roughly~\unit[280]{Hz} and~\unit[5]{Hz}, respectively.
This also shows that the~\glspl{cnn} can speed up the evaluation of the \gls{vfa} and the \gls{vpa} 
up to 4 times and 2 times, respectively.

Note that it takes longer to compute the~\gls{vfa} of this work versus our previous work~\cite{Villarreal2019}.
This is because the~\gls{vfa} of this work considers more inputs than in our previous work, 
and thus, the size of the~\gls{cnn} is larger. 
As can be seen, the~\gls{vpa} runs at a relatively lower update rate compared to the~\gls{vfa}.
We believe that this is not an issue since 
the~\gls{vfa} runs at the legs-level while the~\gls{vpa} runs as the body-level
which means that the legs experience faster dynamics than the body.

During simulations and experiments, 
the~\glspl{cnn} were running on a CPU.
A significant amount of computational time can be reduced if we run the~\glspl{cnn} of the~\gls{vfa} and the~\gls{vpa} on a GPU.
Likewise, a significant amount of computational time can be reduced if a different pose optimization solver is used. 
However, both suggestions are beyond the scope of this work, and are left as a future work. 

\printglossary[title={Abbreviations}]

\section*{Acknowledgments}
The authors would like to thank Geoff Fink and Chundri Boelens for the help provided in this work.
\small
\bibliographystyle{IEEEtran}
\bibliography{./includes/bibliography.bib}
\small

\begin{IEEEbiography}%
[{\includegraphics[width=1in,height=1.28in,clip,keepaspectratio]{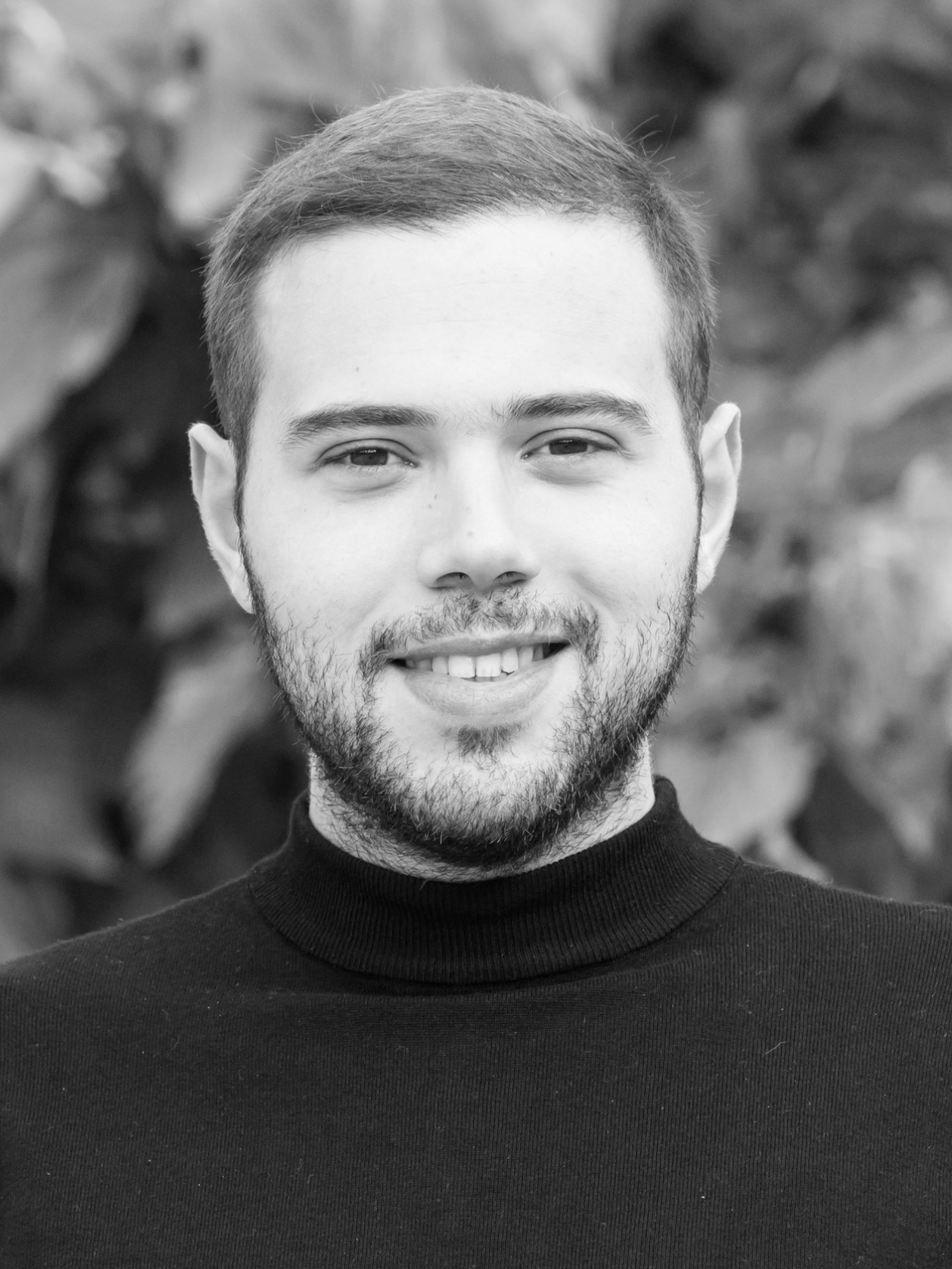}}]%
{Shamel Fahmi}
(S'19) was born in Cairo, Egypt. 
He received the B.Sc. degree in Mechatronics from the German University in Cairo, Cairo, Egypt, in 2015, 
the M.Sc. degree in Systems and Control from the University of Twente, Enschede, the Netherlands, in 2017, and
the Ph.D. degree in Advanced and Humanoid Robotics from the Italian Institute of Technology, Genoa, Italy, in 2021.
His Ph.D. was on terrain-aware locomotion for legged robots. 
Currently, he is researcher at the Biomimetic Robotics Lab, Massachusetts Institute of Technology, MA,~USA.
His research interests include locomotion planning and control for quadruped and humanoid robots using model- and learning-based methods.
\end{IEEEbiography}
~
\begin{IEEEbiography}%
[{\includegraphics[width=1in,height=1.25in,clip,keepaspectratio]{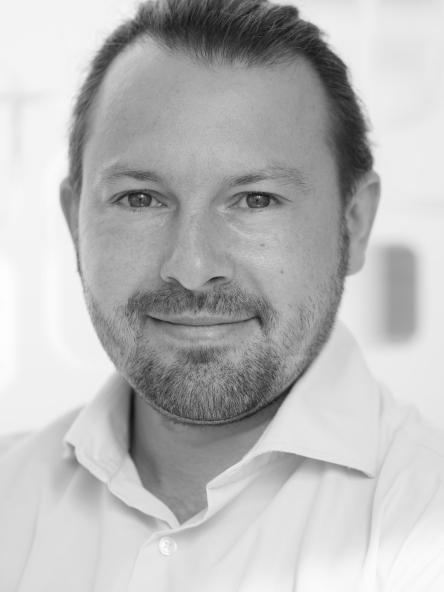}}]%
{Victor Barasuol}
received the Diploma in electrical engineering from Universidade do Estado de Santa Catarina (UDESC), Joinville, Brazil, in 2006.
He has a M.Sc. degree in electrical engineering and a Ph.D. degree in automation and systems engineering, both obtained from Universidade Federal de Santa Catarina (UFSC), Florian\'{o}polis, Brazil, in 2008 and 2013, respectively. He is currently a researcher at the Dynamic Legged Systems (DLS) lab at Istituto Italiano di Tecnologia (IIT), Genoa, Italy. He is knowledgeable from the kinematic design to the control of hydraulic quadruped robots, with major expertise in dynamic motion generation and control. His research activities focus on proprioceptive-based and exteroceptive-based reactions, whole-body control, machine learning for locomotion, active impedance modulation, and loco-manipulation.
\end{IEEEbiography}
~
\begin{IEEEbiography}%
[{\includegraphics[width=1in,height=1.25in,clip,keepaspectratio]{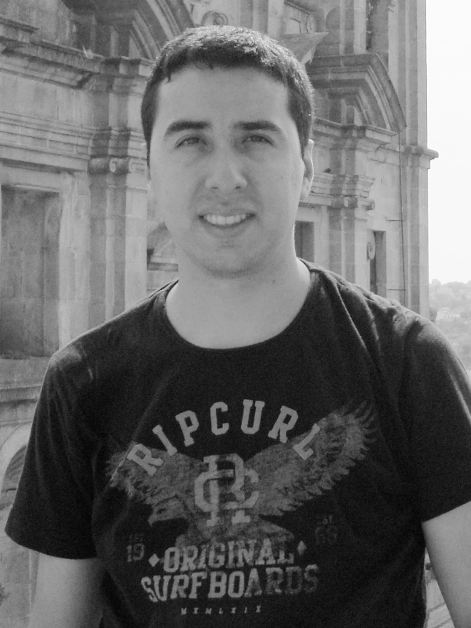}}]
{Domingo Esteban}
received the B.Sc. degree in industrial engineering from the Universidad Nacional de San Agustin, Arequipa, Peru, in 2009, 
the M.Sc. degree in robotics and automation from the Universidad Carlos III, Madrid, Spain, in 2014, 
and the Ph.D. degree in advanced and humanoid robotics from the Italian Institute of Technology (IIT), Genoa, Italy, in 2019.
From 2019 to 2021, he was a Postdoctoral Researcher with the Dynamic Legged Systems Lab at IIT. 
He is currently working at ANYbotics A.G., Zurich, Switzerland. He is interested in machine learning for decision-making and motion control in robotics.
\end{IEEEbiography}
~
\begin{IEEEbiography}%
[{\includegraphics[width=1in,height=1.25in,clip,keepaspectratio]{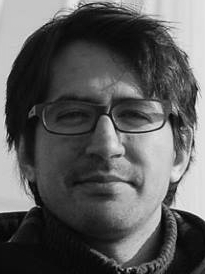}}]
{Octavio Villarreal}
received the M.Sc. degree in mechanical engineering track control engineering from TU Delft, The Netherlands, in 2016, and the Ph.D. degree from IIT and the University of Genoa, in 2019, working at the Dynamic Legged Systems (DLS) lab for his work on vision-based foothold adaptation and locomotion strategies for legged robots. He is currently a robotics research engineer at Dyson Technology Ltd.
\end{IEEEbiography}
~
\begin{IEEEbiography}%
[{\includegraphics[width=1in,height=1.25in,clip,keepaspectratio]{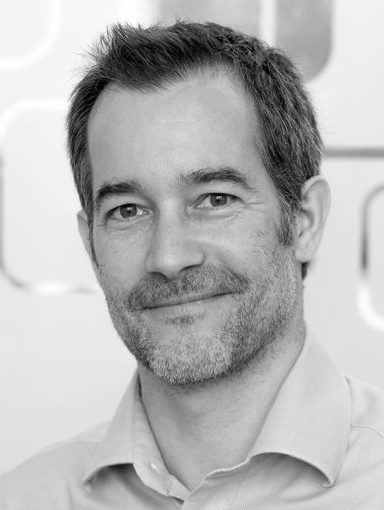}}]%
{Claudio Semini}
(S’07-M’10) received the M.Sc. degree in electrical engineering and information technology from ETH, Zurich, Switzerland, in 2005, and the Ph.D. degree in humanoid technologies from Istituto Italiano di Tecnologia (IIT), Genoa, Italy, in 2010. During his doctorate, he developed the hydraulic quadruped robot HyQ and worked on its control. Since 2012, he leads the Dynamic Legged Systems (DLS) lab. He is a co-founder of the Technical Committee on Mechanisms and Design of the IEEE-RAS Society. He is/was the coordinator/partner of several EU-, National and Industrial projects (including HyQ-REAL, INAIL Teleop, Moog@IIT joint lab, ESA-ANT, etc). His research interests include the design and control of highly dynamic and versatile legged robots for real-world operations, locomotion, hydraulics, and others.
\end{IEEEbiography}

\end{document}